\documentclass[preprint,authoryear,review]{elsarticle} 

\usepackage{graphicx}

\usepackage{amssymb}

\usepackage[top=3.5cm,left=4cm,right=4cm,bottom=3.5cm]{geometry}

\usepackage{afterpage}

\usepackage{enumitem} 
  \setlist{itemsep=1ex plus0.2ex, leftmargin=*, align=left}

\makeatletter
\newcommand{\labitem}[2]{%
\def\@itemlabel{\textbf{#1}}
\item
\def\@currentlabel{#1}\label{#2}}
\makeatother

\makeatletter
\newcommand{\headingitem}[1]{%
\vspace{0.3cm}
\def\@itemlabel{\textbf{#1}}
\item
\def\@currentlabel{#1}
\addtocounter{enumi}{-1}
}
\makeatother

\usepackage{csvsimple}
\usepackage{multirow}
\usepackage{lipsum}

\usepackage{eurosym}
\usepackage{siunitx}

    \DeclareSIUnit\eur{\officialeuro}
    \DeclareSIUnit\M{M}
    \DeclareSIUnit\k{k}

  \def\sym#1{\ifmmode^{#1}\else\(^{#1}\)\fi}

\usepackage{etoolbox}
\robustify\bfseries

\usepackage{subcaption}

\usepackage{setspace} 
\usepackage{csquotes}
\usepackage{amsmath}
\usepackage{bm}

\usepackage{xspace}
	\newcommand\ie{i.\,e.\xspace}
	\newcommand\eg{e.\,g.\xspace}

\usepackage[amsmath,hyperref,framed]{ntheorem} 
  \theoremstyle{plain}
     
  \theoremstyle{nonumberplain}
    \theoremseparator{.}
    \theoremheaderfont{\bfseries}
    \theorembodyfont{\normalfont}
    \theoremsymbol{$\blacksquare$}
      \RequirePackage{amssymb}

      \makeatletter
    \let\copy@theorem@headerfont=\theorem@headerfont
    \newcommand{\my@theorem@headerfont}{%
        \boldmath\copy@theorem@headerfont\unboldmath
      }
    \let\theorem@headerfont=\my@theorem@headerfont
      \makeatother

\theoremstyle{nonumberplain}
\theoremseparator{.}
\newcommand{\argmax}{\operatornamewithlimits{arg \, max}}

\usepackage[%
  sort&compress
]{cleveref}
\usepackage{url}

\usepackage{isomath}

  \newcommand{\abs}[1]{\left\lvert #1 \right\rvert}

\usepackage{algorithm}
\usepackage{algpseudocode}

\usepackage[usenames,dvipsnames]{xcolor}

\usepackage{colortbl}
\usepackage{tabularx}
\usepackage{booktabs}
\usepackage{collcell}

\usepackage{ragged2e}
  
\newcommand{\PreserveBackslash}[1]{\let\temp=\\#1\let\\=\temp}
\newcolumntype{v}[1]{>{\PreserveBackslash\RaggedRight\hspace{0pt}}p{#1}}

  \usepackage{adjustbox}
\newcolumntype{Q}[2]{%
    >{\adjustbox{angle=#1,lap=\width-(#2)}\bgroup}%
    l%
    <{\egroup}%
}

\usepackage{pdflscape}

\newcommand{\mcellt}[2][c]{%
  \begin{tabular}[t]{@{}#1@{}}#2\end{tabular}}

\usepackage[
    colorinlistoftodos,
    textsize=footnotesize,
        ]{todonotes}

\newcommand{\cmark}{\ding{51}}%
\newcommand{\xmark}{\ding{55}}%

\makeatletter
    \renewcommand{\fps@figure}{htb}         
    \renewcommand{\fps@table}{H}         
\makeatother 

\usepackage{float}
\usepackage{rotating}

\journal{Expert Systems with Applications}

\begin{document}

\begin{frontmatter}




\title{Sentiment analysis based on rhetorical structure theory:\\Learning deep neural networks from discourse trees}


\author[ETH]{Mathias Kraus\corref{cor1}}
\ead{mathiaskraus@ethz.ch}

\author[ETH]{Stefan Feuerriegel}
\ead{sfeuerriegel@ethz.ch}

\address[ETH]{ETH Zurich, Weinbergstr. 56/58, 8092 Zurich, Switzerland}
\cortext[cor1]{Corresponding author.}

\begin{abstract}
Prominent applications of sentiment analysis are countless, covering areas such as marketing, customer service and communication. The conventional bag-of-words approach for measuring sentiment merely counts term frequencies; however, it neglects the position of the terms within the discourse. As a remedy, we develop a discourse-aware method that builds upon the discourse structure of documents. For this purpose, we utilize rhetorical structure theory to label \mbox{(sub-)clauses} according to their hierarchical relationships and then assign polarity scores to individual leaves. To learn from the resulting rhetorical structure, we propose a tensor-based, tree-structured deep neural network (named \mbox{Discourse-LSTM}) in order to process the complete discourse tree. The underlying tensors infer the salient passages of narrative materials. In addition, we suggest two algorithms for data augmentation (node reordering and artificial leaf insertion) that increase our training set and reduce overfitting. Our benchmarks demonstrate the superior performance of our approach. Moreover, our tensor structure reveals the salient text passages and thereby provides explanatory insights.
\end{abstract}

\begin{keyword}
Sentiment analysis \sep Rhetorical structure theory \sep Discourse tree \sep Tree-structured network \sep Long short-term memory \sep Tensor-based network\\[0.3cm]
\noindent \textbf{Declarations of interest:} \emph{none}

\end{keyword}

\end{frontmatter}



\section{Introduction}
\label{sec:Introduction}

Sentiment analysis reveals personal opinions towards entities such as products, services or events, which can benefit organizations and businesses in improving their marketing, communication, production and procurement. For this purpose, sentiment analysis quantifies the positivity or negativity of subjective information in narrative materials \citep{Pang.2008,Feldman.2013,Chen.2017,Kratzwald.2018}. Among the many applications of sentiment analysis are tracking customer opinions \citep{Tanaka.2010,Araque.2017,Bohanec.2017}, 
 mining user reviews \citep{Ye.2009,Mostafa.2013,Kontopoulos.2013}, trading upon financial news \citep{KhadjehNassirtoussi.2015,Kraus.2017,Weng.2018}, detect social events \citep{Yoo.2018} and predicting sales \citep{Yu.2012,Rui.2013}.

Sentiment analysis traditionally utilizes bag-of-words approaches, which merely count the frequency of words (and tuples thereof) to obtain a mathematical representation of documents in matrix form \citep{Manning.1999,Pang.2008,Guzella.2009,Dey.2018}. As such, these approaches are not capable of taking into consideration semantic relationships between sections and sentences of a document. In na{\"i}ve bag-of-words models, all clauses are assigned the same level of relevance, which cannot mark certain subordinate clauses more than others for purposes of inferring the sentiment. Conversely, the objective of this paper is to develop a discourse-aware method for sentiment analysis that can recognize differences in salience between individual subordinate clauses, as well as the discriminate the relevance of sentences based on their function (\eg whether it introduces a new fact or elaborates upon an existing one).

\begin{figure}
\centering
\makebox[\textwidth]{%
\begin{tabular}{cc}
(a) Discourse with overall positive sentiment & (b) Discourse with overall negative sentiment \\
\includegraphics[width=0.45\linewidth]{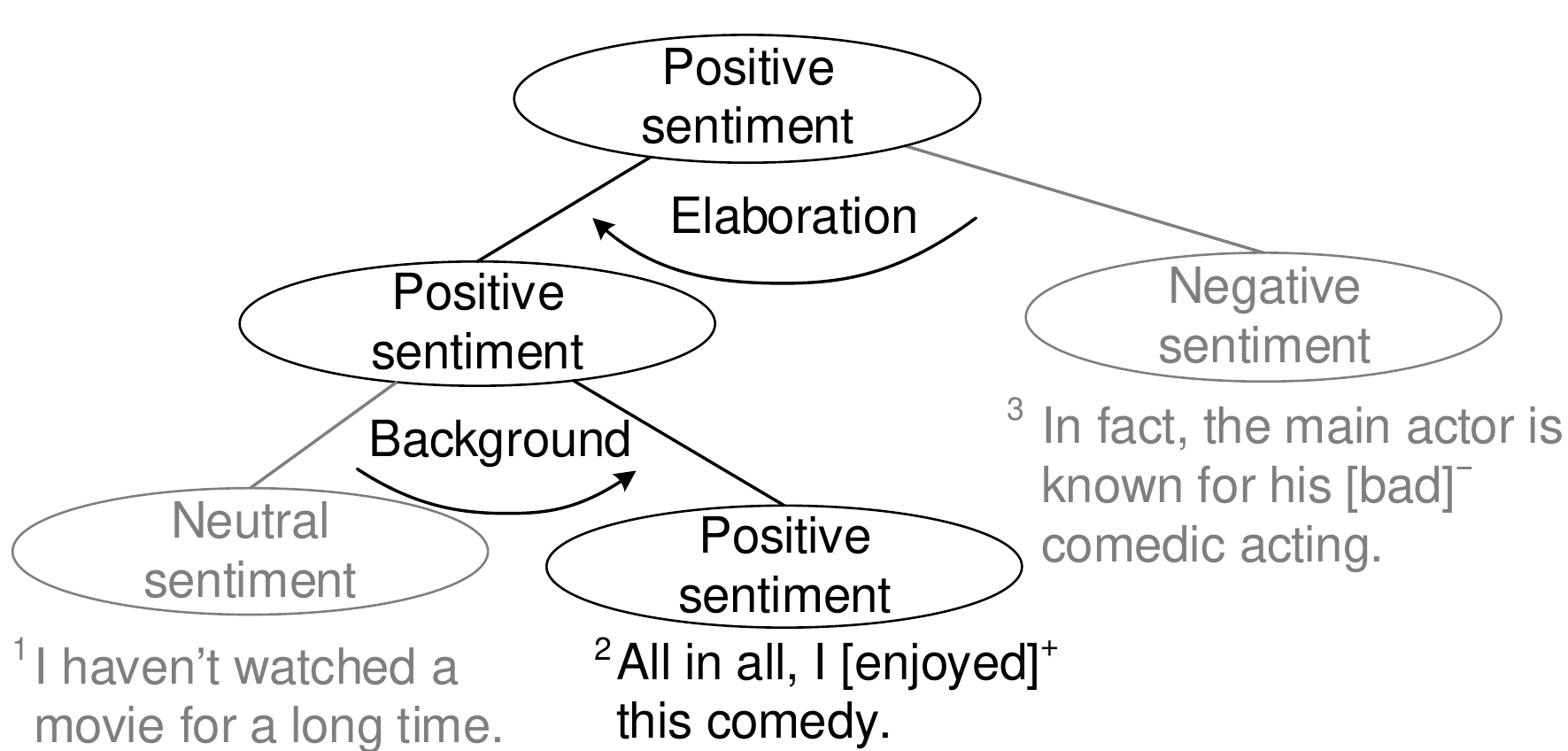} & \includegraphics[width=0.45\linewidth]{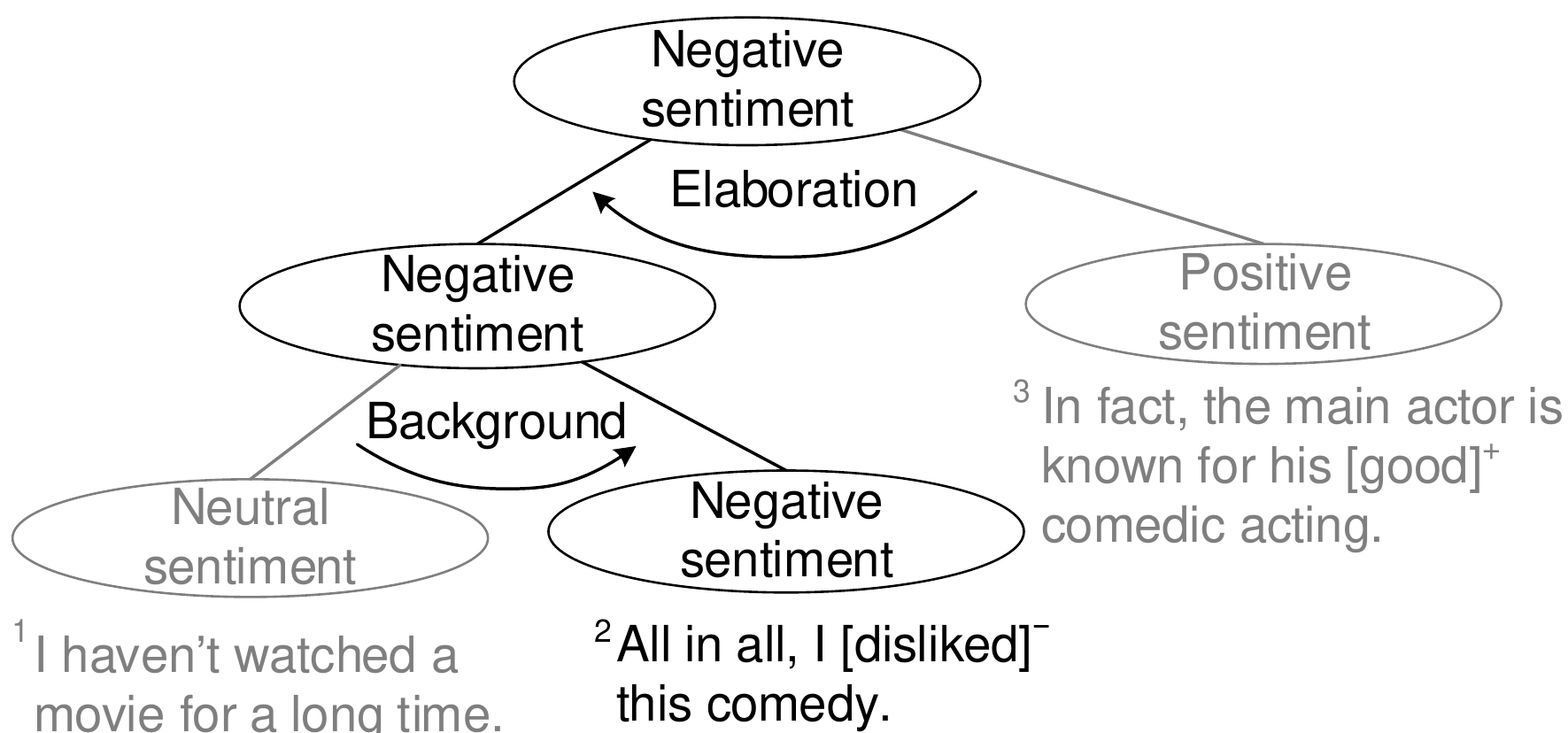} \\
\end{tabular}
}%
\caption{Illustrative examples in which the discourse tree helps identify the conveyed sentiment from the main clause (highlighted in black). Here relation type additionally denotes the rhetorical function. The original inputs are: \emph{\textquote{I haven't watched a movie for a long time. All in all, I liked/disliked this comedy. In fact, the main actor is known for his bad/good comedic acting.}}.}
\label{fig:RST_tree_sentiment}
\end{figure}

Let us, for instance, consider the two examples in \Cref{fig:RST_tree_sentiment}, which express opposite polarities. By simply counting the frequency of positive and negative words, we cannot discriminate between the texts, as both contain the same number of polarity terms. To reliably analyze the sentiment, it is essential to account for the semantic structure and the variable importance across passages. That is, we can identify the main clauses and then infer the correct tone of the examples by looking at them. Similarly, RST trees can locate relevant parts in lengthy texts. For instance, the concluding section of a newspaper article is typically relevant as it reports the opinion of the author.

Our method is based on rhetorical structure theory~(RST), which incorporates the discourse structures of natural language. RST structures documents hierarchically \citep{Mann.1988} by splitting the content into \mbox{(sub-)clauses} called elementary discourse units~(EDUs). The EDUs are then connected to form a binary discourse tree. Here RST discriminates between a nucleus, which conveys primary, and satellite, which conveys ancillary information. The formalization of nucleus/satellite can be loosely thought of main and subordinate parts of a clause. The edges are further labeled according to the type of discourse -- for instance, whether it is an elaboration or an argument. Hence, this method essentially derives the function of a text passage. 
 Both concepts of the RST tree help in localizing essential information within documents. Hence, the goal of this work is to develop a novel approach that identifies salient passages in a document based on their position in the discourse tree and incorporates their importance in the form of weights when computing sentiment scores.

Previous research has demonstrated that discourse-related information can improve the performance of sentiment analysis (see \Cref{sec:background} for details). The work by \citet{Taboada.2008} is the first to combine rhetorical structure theory and sentiment analysis. In this work, the authors weigh adjectives in a nucleus more heavily than those in a satellite. Beyond that, one can reweigh the importance of passages based on their relation type \citep{Hogenboom.2015} or depth \citep{Markle.2017} in the discourse tree. Some methods prune the discourse trees at certain thresholds to yield a tree of fixed depth, \eg \num{2} or \num{4} levels \citep{Markle.2017}. Other approaches train machine learning classifiers based on the relation types as input features \citep{Hogenboom.2015b}. What the previous references have in common is that they try to map the tree structure onto mathematically simpler representations, thereby dropping partial information from the tree. 

An alternative strategy is to apply tree-structured neural networks that traverse discourse trees for representation learning. When encountering a node, these networks combine the information from the leaves and pass them on to the next higher level, until reaching the root at which point a prediction is made. Thereby, the approach merely adheres to the tree-structure but does not account for either the relation type or whether it is a nucleus/satellite. To do so, one can extend the network to include different weights for each edge in the tree depending on, \eg, the relation type. This essentially introduces additional degrees of freedom that can weigh the different discourse units by their importance. The work by \citet{Fu.2016} extends the network by such a mechanism with respect to the nucleus/satellite information but discards the relation type and merely applies the network to individual sentences instead of longer documents. The approach in \citet{Ji.2017} can only exploit the relation type and not the nucleus/satellite information. Furthermore, former approaches are based on traditional recursive neural networks, which are limited by the fact that they can persist information for only a few iterations \citep{Bengio.1994}. Therefore, these methods struggle with complex discourses, while we explicitly build upon tree-shaped long short-term memory models, since they are better equipped to handle very deep structures.

We build upon the previous works and advance them by proposing a specific neural network, called \emph{Discourse-LSTM}. The Discourse-LSTM utilizes multiple tensors to localize salient passages within documents by incorporating the full discourse structure including nucleus/satellite information \emph{and} relation types. In brief, our approach is as follows: we utilize rhetorical structure theory to represent the semantic structure of a document in the form of a hierarchical discourse tree. We then obtain sentiment scores for each leaf by utilizing both polarity dictionaries and word embeddings. The resulting tree is subsequently traversed by the Discourse-LSTM, thereby aggregating the sentiment scores based on the discourse structure in order to compute a sentiment score for the document. This approach thus weighs the importance of \mbox{(sub-)clauses} based on their position and relation in the discourse tree, which is learned during the training phase. As a consequence, this allows us to enhance sentiment analysis with discourse information. Another key contribution is that we propose two techniques for data augmentation that facilitate training and yield higher predictive accuracy.

The remainder of this paper is structured as follows. \Cref{sec:background} reviews discourse parsing and RST-based sentiment analysis. \Cref{sec:method} then introduces our Discourse-LSTM, as well as our algorithms for data augmentation. \Cref{sec:experiment} describes our experimental setup in order to evaluate the performance of our deep learning methods in comparison to common baselines (\Cref{sec:results}). \Cref{sec:conclusion} concludes with a summary and suggestions for future research.

\section{Background}
\label{sec:background}

\subsection{Rhetorical structure theory}
\label{sec:RST}

\begin{figure}
\centering
\includegraphics[width=.6\linewidth]{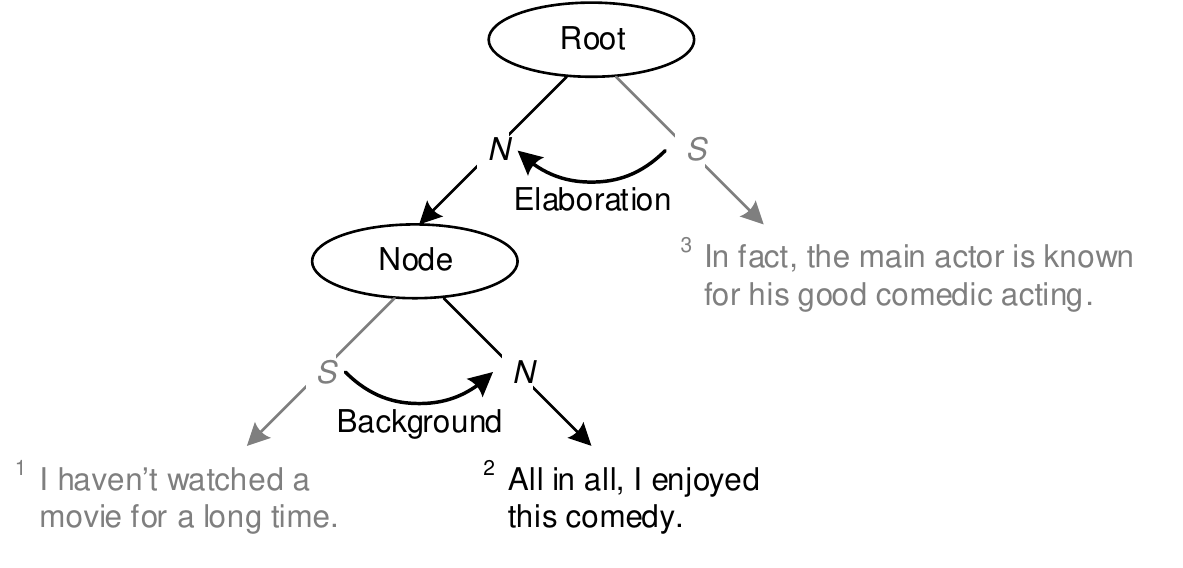}
\caption{Example discourse tree with \num{3} elementary discourse units, for which $N$ denotes a nucleus and $S$ a satellite.}
\label{fig:RST_Tree}
\end{figure}

Rhetorical structure theory formalizes the discourse in narrative materials by organizing \mbox{sub-clauses}, sentences and paragraphs into a hierarchy \citep{Mann.1988}. The premise is that a document is split into elementary discourse units, which constitute the smallest, indivisible segments. These EDUs are then connected by one of \num{18} different relation types, which represent edges in the discourse tree; see \Cref{tbl:rst_relations} for a list. Each relation is further labeled by a hierarchy type, \ie either as a nucleus~($N$) or a satellite~($S$). Here a nucleus denotes a more essential unit of information, while a satellite indicates a supporting or background unit of information. We note that RST also defines cases where both children are labeled as nuclei at the same time. \Cref{fig:RST_Tree} presents an example of a discourse tree. Here the label \emph{elaboration} at the root indicates that sentence~3 provides an additional detail about the content (\ie the comedy) of the left sub-tree. Furthermore, \emph{background} reveals that sentence~1 increases the comprehensibility of sentence~3, since it is needed to make sense of the phrase \emph{\textquote{all in all}}.

\begin{table}
\footnotesize
\begin{tabular}{l l}
\toprule
{\textbf{Relation type}} & {\textbf{Description}} \\
\midrule 
Elaboration & Satellite provides additional details about the nucleus \\
Joint & No specific hierarchy between EDUs \\
Same-unit & Links parts of one EDU to another \\
Background & Satellite provides information to comprehend nucleus \\
Attribution & Satellite contains reporting verbs or cognitive predicates for nucleus \\
Comparison & Refers to similarities and dissimilarities \\
Temporal & Describes a specific ordering in time between units \\
Enablement & Satellite increases the ability to perform the action in nucleus \\
Contrast & Describes comparability or differences \\
Summary & Satellite is a shorter restatement of nucleus  \\
Condition & Realization of nucleus depends on realization of satellite \\
Manner-means & Satellite tends to make realization of nucleus more likely \\
Cause & Satellite is a reason for nucleus \\
Explanation & Satellite justifies information from nucleus \\
Evaluation & Satellite assesses nucleus \\
Textual-organization & Describes the composition of the document \\
Topic-change & Topic has changed between units \\
Topic-comment & One EDU annotates another \\
\bottomrule
\end{tabular}
\caption{Overview of the different relation types that connect elementary discourse units \citep{Mann.1988}.}
\label{tbl:rst_relations}
\end{table}

Previous research has proposed various methods for automating the parsing of discourse trees of documents. Common implementations for documents consisting of multiple paragraphs are represented by the high-level discourse analyzer HILDA \citep{Hernault.2010} and the DPLP parser \citep{Ji.2014}, of which the DPLP parser currently achieves the better F1-score in identifying relation types. Although DPLP is slightly outperformed by HILDA in EDU span detection by \SI{1.4}{\percent} in terms of the F1-score, it shows an improvement of \SI{2.6}{\percent} and \SI{6.9}{\percent} on identifying the hierarchy and relation types, respectively \citep{Ji.2014}. Since inferring relation types is regarded as the most challenging subtask of RST parsing, we decided to utilize the DPLP parser in this work.

Besides RST, other forms of semantic representations have also been devised \citep{Abend.2017}. These include logical structures which put a focus on quantifications, negations and coordination, while other works involve temporal relations, inferences and textual entailment. Further frameworks are speech-act theory and natural semantic metalanguage. However, their labeling is most often not unique and applied merely at sentence level without hierarchical structures. In contrast, RST specifically entails characteristics that provide benefits in our case: we obtain a hierarchical and fully-connected representation that covers the complete document \citep{Liu.2018b}. 
 Accordingly, our methodology adapts to the underlying story, discriminating between key messages and supplementary materials.

\subsection{Sentiment analysis with RST}

Previous studies have advocated different approaches for sentiment analysis that utilize the discourse tree. In the following, we categorize these approaches into (a)~weighting rules or (b)~tree-structured neural networks.   


The paper by \citet{Taboada.2008} is the first work that explicitly utilizes rhetorical structure theory in order to extract sentiment from linguistic content. It determines the relevance of words depending on whether they appear in a nucleus or satellite. Subsequently, further works have developed different weighting rules (see \Cref{tbl:literature_weighting}). These aggregate the sentiment scores of EDUs based on the tree structure \citep{Heerschop.2011,Hogenboom.2015}. However, the weights are frequently pre-determined and hand-crafted. A different stream of research also considers hierarchy labels (nucleus or satellite) of the nodes and updates the weights based on these. Examples include approaches that focus on the top-split (\ie the root node) of the discourse tree and scale the relative importance based on (hand-crafted) weights \citep{Taboada.2008, Heerschop.2011, Hogenboom.2015}. The underlying weights can also be optimized using logistic regression \citep{Chenlo.2014}. Hierarchy labels at leaf level also facilitate a more fine-grained evaluation \citep{Hogenboom.2015}, even though the discourse tree from above is neglected. Recent research also applies a recursive weighting scheme that utilizes a scaling factor to reduce the influence of passages from lower parts of the discourse tree \citep{Hogenboom.2015, Markle.2017}. Alternatively, one can prune the discourse tree at certain thresholds in order to yield a tree of fixed depth, \eg \num{2} or \num{4} levels \citep{Markle.2017}. Some works also incorporate relation types between EDUs \citep{Heerschop.2011, Chenlo.2014, Hogenboom.2015} or categorize them into contrastive or non-contrastive relations, which are then weighted separately \citep{Zirn.2011}. What the previous rule-based approaches have in common is that they cannot incorporate the complete tree into their analysis and, instead, need to partially discard discourse information, \ie the links between nodes within the tree structure.


\Cref{tbl:literature_TreeLSTM} provides an overview of papers utilizing tree-structured approaches. The RST tree can be traversed with a recursive neural network \citep{Bhatia.2015}; however, this approach only incorporates the relation types and lacks information regarding the hierarchy type. The work by \citet{Fu.2016} applies a Tree-LSTM to the discourse trees and extends this method to discriminate nucleus and satellite but at the same time neither discerning the relation type nor applying data augmentation. A similar approach traverses the RST tree with the help of a recursive neural network, while utilizing relation-specific composition matrices \citep{Ji.2017}. However, the recursive neural network is known to struggle with complex tree structures because of vanishing or exploding gradients and, instead, we utilize a long short-term memory. Moreover, the approach sums the representations in each recursion and, hence, cannot distinguish the hierarchy, \ie between nucleus and satellite. Hence, the objective of this paper is to extend the previous works by advancing representation learning in order to incorporate the complete discourse tree, including relation types, tree depth and hierarchy labels.

\afterpage{%
\newgeometry{left=1cm,right=1cm,bottom=3.5cm}
\begin{landscape}
\thispagestyle{empty}
\begin{minipage}{\linewidth}
\thispagestyle{empty}
\begin{table}[H]
\tiny
    \begin{tabular}{ll lll ccc}
		    \toprule
		    {\textbf{Reference}} 					& {\textbf{Weight optimizations}} & {\textbf{EDU features}} & {\textbf{RST parser}}	 & {\textbf{Model}} & \multicolumn{3}{c}{\textbf{Considered RST features}} \\
		    \cmidrule(l){6-8} 
										& 					    & 				      & 			              & 			      & {\textbf{Relation}} & {\textbf{Tree}} & {\textbf{Nucleus/}}  \\
		    & & & & & {\textbf{type}} & {\textbf{depth}} & {\textbf{satellite}} \\
		    \midrule					   
 		    \citet{Chenlo.2014} 	&Weights optimized 		& Dictionary-based 		& SPADE	 			&  Top-split weighting	 & \cmark & \xmark & \cmark  \\
		    &  by logistic regression		 & sentiment score 		&					 &  & &   \\[0.5cm]
		    \citet{Heerschop.2011}	&  Handcrafted weighting factor	 & Dictionary-based 	& SPADE				& Position-based weighting rule & \xmark & \xmark & \xmark \\
		    \cline{5-8}
		   								 & or optimized by			 & sentiment score 		& 					& Top-split weighting rule         	& \xmark & Up to level \num{1} & \cmark \\
		    \cline{5-8}
		    							& genetic algorithm			 & 				& 					&  Relation-type weighting 	& \cmark & \xmark & \cmark  \\[0.5cm]
		    \citet{Hogenboom.2015} &  Handcrafted weighting factor	 & Dictionary-based 	& HILDA				& Position-based weighting	 & \xmark & \xmark & \xmark  \\
		    \cline{5-8}
		   								 &  or particle swarm optimization	& sentiment score 		& SPADE				& Top-split weighting		& \xmark &  Up to level \num{1} & \cmark  \\
		    \cline{5-8}
		    								&  					& 				& 					& Bottom-split weighting		& \xmark & \xmark & \cmark  \\
		    \cline{5-8}
		    								& 				 	& 				& 					& Hierarchical weighting 		& \xmark & \cmark & \cmark  \\[0.5cm]
		    \citet{Hogenboom.2015b} & Weights optimized by SVM	 & Dictionary-based		 &  SPADE				& Top- and leaf-split weighting	& \cmark & \xmark & \xmark  \\
		    								& 				 	& sentiment score		 &		 			& of \num{14} relation types \\[0.5cm]
		    \citet{Markle.2017} 	& Weights optimized by grid-search  & Dictionary-based 	& HILDA				&  Hierarchical weighting rule	 & \xmark & \cmark & \cmark  \\ 	    
		    \cline{5-8}
		    								&					 & sentiment score 		& 					&  Random forest on pruned tree 	& \xmark & Up to level & \cmark   \\
		    								& 					& 				& 					& 					& 	     & \num{2} or \num{4} &   \\[0.5cm]
 		   \citet{Taboada.2008} & Manually-chosen weights		& Adjective-based 		&   SPADE			& Different weighting of 		& \xmark & Up to level \num{1} & \cmark  \\
		    								& 					& sentiment score 		& 					& nucleus vs. satellite of top-split  \\
       \bottomrule
\end{tabular}
\caption{Comparison of methods for sentiment analysis proposing weighting schemes that utilize the discourse structure.}
\label{tbl:literature_weighting}
\end{table}%

\begin{table}[H]
\tiny
    \begin{tabular}{l p{3cm}ll p{3cm} ccc p{2.5cm}}
		    \toprule
		    {\textbf{Reference}} 					& {\textbf{Model}} & {\textbf{Tensor}} & {\textbf{EDU features}} & {\textbf{RST parser}}	 & \multicolumn{3}{c}{\textbf{Considered RST features}} & {\textbf{Additional characteristics}} \\
		    \cmidrule(l){6-8} 
										& 					    &  {\textbf{structure}} & 				      & 			              & {\textbf{Relation}} & {\textbf{Tree}} & {\textbf{Nucleus/}}  \\
		    & & & & {\textbf{type}} & {\textbf{depth}} & {\textbf{satellite}} \\
		    \midrule					   
		    \citet{Bhatia.2015} 	& Rhetorical recursive neural networks & Relation type & Dictionary-based & DPLP & \xmark & \cmark & \cmark \\
				&  with recurrent neural network & & sentiment scores & &   \\[0.5cm]
				\citet{Fu.2016} & Tree-LSTM ($2$-ary)	& Nucleus/satellite & Linear combination	& Dependency and			 & \xmark & \cmark & \cmark & Only single sentences as input \\
				& & & of word embeddings & constituent parser \citep{Surdeanu.2015}  \\[0.5cm]
		    \citet{Ji.2017} 			&  Recursive neural network  & Relation type	& Hidden states of 		& DPLP  & \cmark & \cmark & \xmark \\
										&  	& & bidirectional LSTM  \\
										& 				 	& & on word embeddings 	  \\
 		    \midrule
 		    {\textbf{This paper:}} 					& Tree-LSTM ($N$-ary, child-sum) 	& Relation type \emph{and} & Dictionary-based 		&  DPLP 		& \cmark & \cmark & \cmark & Procedures for data augmentation     \\ 
 		    							\textbf{Discourse-LSTM}	&  	& nucleus/satellite	& sentiment scores, 		&					 \\
 		    								&  	& & word embeddings 		&					&  \\
        \bottomrule
\end{tabular}
\caption{Comparison of methods for sentiment analysis proposing tree-structured approaches based on neural networks.}
\label{tbl:literature_TreeLSTM}
\end{table}
\end{minipage}
\end{landscape}
\clearpage
\restoregeometry
}


The features used by the aforementioned papers differ. On the one hand, sentiment scores for EDUs are computed from dictionaries (where words are labeled as positive or negative). In terms of dictionaries, common examples include SentiWordNet \citep{Zirn.2011, Heerschop.2011, Hogenboom.2015, Hogenboom.2015b,Liu.2018}, hand-crafted dictionaries \citep{Taboada.2008} or domain-specific dictionaries \citep{Markle.2017}. On the other hand, approaches utilize vector representations for the EDUs based on word embeddings \citep{Fu.2016,Ji.2017}. For reasons of comparability, we also utilize both a dictionary-based approach and word embeddings in order to compute sentiment features from the content of elementary discourse units.

\subsection{Representation learning for sequential and tree data}

Recent advances in deep neural networks have rendered it possible to learn representations of unstructured data such as sequences, texts or trees \citep{Goodfellow.2017}. This can, for instance, be achieved by recurrent neural networks, which entail an internal architecture in the form of a directed cycle, thereby creating an internal state encoding dependent structures \citep{Chen.2017}. 
 Based on these, one can process texts of arbitrary length in sequential order, while the internal state learns the complete sequence and passes information from one word to the next. However, in practice, information only persists for a few iterations \citep{Bengio.1994}. A viable remedy is provided by the long short-term memory~(LSTM) network. The LSTM enhances recurrent neural networks by capturing long dependencies among input signals \citep{Hochreiter.1997}. 


Previous research has proposed a Tree-LSTM that can deal with representation learning for trees. This tree-structured LSTM network traverses trees bottom-up in order to generate representations of the underlying structure \citep{Tai.2015}. The Tree-LSTM computes a representation for each parent node based on its immediate children and does so recursively until the root of the tree is reached. It thereby stacks individual LSTMs such that they reflect the tree structure from the input. However, the Tree-LSTM provides no possibility of incorporating additional information from the discourse trees, such as the relation type. The Tree-LSTM can be applied to RST trees and we thus rely upon it as a baseline. We later extend the na{\"i}ve Tree-LSTM through two tensor structures that express the additional degrees of freedom. This results in a Discourse-LSTM that allows us to utilize the complete set of information encoded in discourse tree. 

\section{Discourse-based sentiment analysis with deep learning}
\label{sec:method}

\begin{figure}
\centering
\includegraphics[width=.7\textwidth]{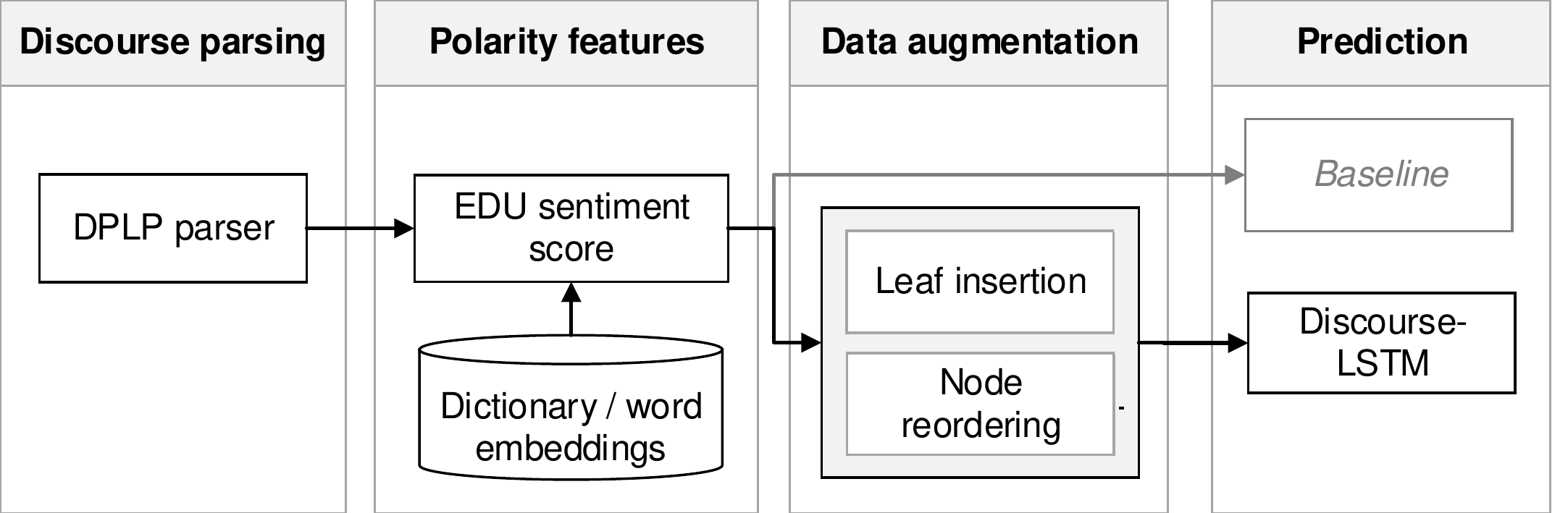}
\caption{Research framework evaluating the gains in predictive performance from combining our Discourse-LSTM and data augmentation in comparison to the baselines.}
\label{fig:framework}
\end{figure}

This section introduces our discourse-based methodology, which infers sentiment scores from textual materials. \Cref{fig:framework} illustrates the underlying framework and divides the procedure into steps for discourse parsing, computing low-level polarity features, data augmentation and prediction. The prediction phase implements either of the baselines or our proposed Discourse-LSTM. 

\subsection{Discourse parsing}
\label{sec:discourse_parsing}

We generate discourse trees for our datasets by utilizing the DPLP parser \citep{Ji.2014}. For sake of simplicity, we introduce the following notation. We denote the relation type of node $i$ as $\rho_i \in \{ \mathrm{elaboration}, \mathrm{argument}, \ldots \}$. The complete list of relation types is given in \Cref{tbl:rst_relations}. Furthermore, we introduce $\tau_i \in \{ \mathrm{nucleus}, \mathrm{satellite} \}$ as the hierarchy type of node $i$.  

\subsection{Polarity features}

We follow common procedures in sentiment analysis and utilize both a pre-defined dictionary that labels terms as positive or negative \citep{Pang.2008,Feldman.2013}, and word embeddings that represent text in multiple dimensions \citep{Fu.2016, Ji.2017}. 

Sentiment dictionaries have multiple advantages, as they are domain-independent and work reliably even with few training observations. In addition, one can easily exchange the underlying dictionary for one that not only measures polarity or negativity, but is concerned with other language concepts such as subjectivity, certainty or the domain-specific tone. Our experimental results are based on the SentiWordNet~3.0 dictionary \citep{Baccianella.2010}, which provides sentiment labels for \num{117659} words. Based on the sentiment labels at word level, we then proceed to compute a sentiment score $\sigma_i$ for each EDU $i$ via 
\begin{equation}
\sigma_i = \frac{1}{\abs{\left\{ w \,|\, w \in i \right\}}}~ \sum_{w \in i} \text{pos}(w) - \text{neg}(w) ,
\label{equ:net-optimism}
\end{equation}
where we iterate over the words $w$ in EDU $i$, while $\text{pos}(w)$ and $\text{neg}(w)$ are the positivity and negativity scores for word $w$ according to SentiWordNet. The resulting sentiment value $\sigma_i$ thus represents the low-level features that later serve as input to our predictive models.

In addition, we utilize a fully neural approach by incorporating multi-dimensional word embeddings which contain considerably more information than sentiment values. In particular, we employ pre-trained \num{50}-dimensional word embeddings from GloVe\footnote{https://nlp.stanford.edu/projects/glove/} to represent words in each EDU. Based on the word representations in each EDU $i$, we form a high-level feature vector $\sigma_i$, representing the EDU, via 
\begin{equation}
\sigma_i = \frac{1}{\abs{\left\{ w \,|\, w \in i \right\}}}~ \sum_{w \in i} e^w_i ,
\label{equ:word_avg}
\end{equation}
with $e^w_i$ being the word embedding of word $w$ in EDU $i$. This approach of forming representations has been shown to work well on short texts, as is the case for RST leaves \citep{Boom.2016}.

\subsection{Tree-LSTM baseline}
\label{sec:TreeLSTM}

We draw upon the Tree-LSTM as a baseline similar to \citet{Fu.2016}, since it is widely regarded as the status quo for tree learning \citep{Tai.2015}. The Tree-LSTM takes a discourse tree as input and then processes EDU features while accounting for their position in the tree. For this purpose, it stacks individual LSTMs in the form of that tree and adapts the ideas of both a memory cell and gates from traditional LSTMs, but extends these concepts to tree structures \citep{Tai.2015}. Here the underlying LSTM helps to overcome the problem of exploding gradients.

\begin{figure}
\centering
\makebox[\textwidth]{%
\includegraphics[width=.8\linewidth]{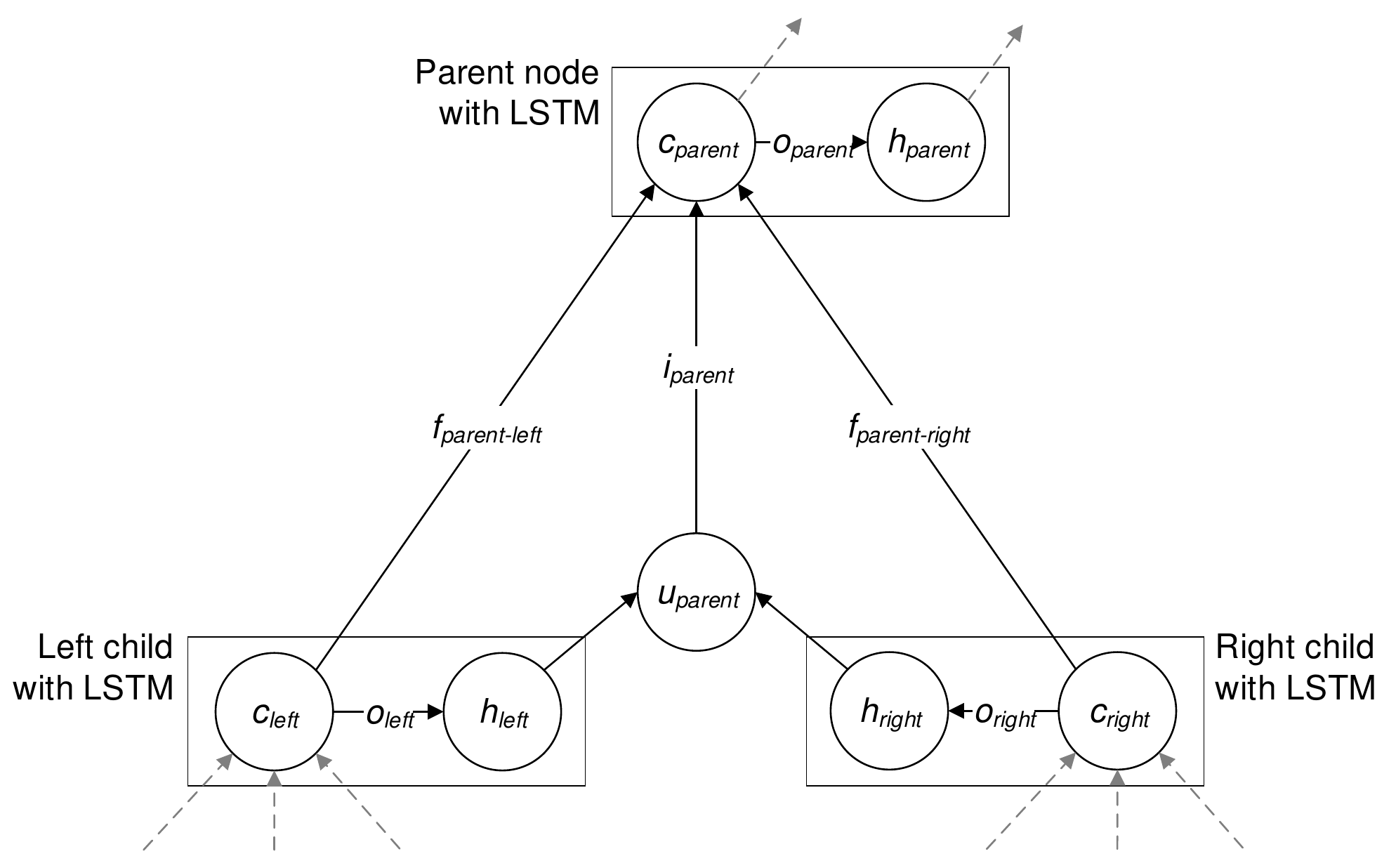}
}
\caption{This schematic illustration shows the composition of the memory state $c_\mathrm{parent}$ and the hidden state $h_\mathrm{parent}$ in a Tree-LSTM with two child nodes.}
\label{fig:tree_lstm_cell}
\end{figure}

In the Tree-LSTM, each node $j$ from the discourse tree is translated into a single LSTM unit, which comprises an input gate $i_j$, an output gate $o_j$, a memory cell $c_j$ and hidden state $h_j$. In contrast to the standard LSTM, the Tree-LSTM contains not a single forget gate, but rather one forget gate $f_{jk}$ for each child $k$. This allows each parent node to recursively compute a representation from its immediate children. The input vectors to each LSTM unit are given by the hidden state $h_k$ and the memory cell $c_k$ for all children $k \in C(j)$, where $C(j)$ is the set of children of parent $j$. This layout of arranging connections renders it possible for the Tree-LSTM to pass information upward in the tree, since every node can incorporate selected information from each child-LSTM. \Cref{fig:tree_lstm_cell} details the connection between the gates in a Tree-LSTM.

Our experiments later compare the performance of two different architectures of Tree-LSTM models, namely, the child-sum and $N$-ary Tree-LSTM \citep{Tai.2015}. Both are common in research, but vary in their connections between input and output gates. The former, the child-sum Tree-LSTM, sums the hidden states $h_k$ from the children $k \in C(j)$ in order to obtain a single input to the hidden state $\tilde{h}_\mathrm{parent}$ of the parent. This approach discards any information regarding the order of the children, since it uses the same weights $U^{(i)}$, $U^{(f)}$, $U^{(o)}$ and $U^{(u)}$ for all children. In contrast, the $N$-ary Tree-LSTM requires a fixed, pre-defined number of $N = \abs{C(j)}$ children for each inner node. It then combines the child nodes by weighting their hidden states based on parameters $U^{(i)}_m$, $U^{(f)}_{km}$, $U^{(o)}_m$ and $U^{(u)}_m$ dependent on the index $m = 1, \ldots, N$ of the child.

Mathematically, for input $x_j \in \mathbb{R}^m$, the child-sum Tree-LSTM transition equations are defined as
\begin{align}
\tilde h_j &= \sum_{k \in C(j)} h_k,\\
i_j &= \mathrm{sigmoid} \left( W^{(i)} x_j + U^{(i)} \tilde h_j + b^{(i)} \right),\\
f_{jk} &= \mathrm{sigmoid} \left( W^{(f)} x_j + U^{(f)} h_k + b^{(f)} \right), \\
o_j &= \mathrm{sigmoid} \left( W^{(o)} x_j + U^{(o)} \tilde h_j + b^{(o)} \right),\\
u_j &= \text{tanh} \left( W^{(u)} x_j + U^{(u)} \tilde h_j + b^{(u)} \right),\\
c_j &= i_j \odot u_j + \sum_{k \in C(j)} f_{jk} \odot c_k,\\
h_j &= o_j \odot \text{tanh}(c_j),
\end{align}
where $\odot$ denotes the element-wise multiplication. Moreover, the above equations contain the weights $W^{(i)}$, $W^{(f)}$, $W^{(o)}$, $W^{(u)}$, each of dimension $n \times n$ for pre-defined memory size $n$, and $b^{(i)}$, $b^{(f)}$, $b^{(o)}$, $b^{(u)}$ of length $n$. Similarly, the $N$-ary Tree-LSTM obtains its memory cell and hidden state via
\begin{align}
i_j &= \mathrm{sigmoid} \left(W^{(i)} x_j + \sum_{m=1}^{N} U_m^{(i)} h_{jm} + b^{(i)}\right),\\
f_{jk} &= \mathrm{sigmoid} \left(W^{(f)} x_j + \sum_{m=1}^{N} U_{km}^{(f)} h_{jm} + b^{(f)}\right), \\
o_j &= \mathrm{sigmoid} \left(W^{(o)} x_j + \sum_{m=1}^{N} U_m^{(o)} h_{jm} + b^{(o)}\right),\\
u_j &= \text{tanh}\left(W^{(u)} x_j + \sum_{m=1}^{N} U_m^{(u)} h_{jm} + b^{(u)}\right),\\
c_j &= i_j \odot u_j + \sum_{m=1}^{N} f_{jm} \odot c_{jm},\\
h_j &= o_j \odot \text{tanh}(c_j).
\end{align}

In order to make sentiment predictions from the Tree-LSTM at the root node, we introduce an additional feedforward classification layer. Here we utilize a softmax classifier that predicts a class label $y$ from the hidden state $h_\mathrm{root}$ of the root node. The softmax layer entails further weights $W^{(s)} \in \mathbb{R}^{n \times n}$ and $b^{(s)} \in \mathbb{R}^n$, based on which it computes the probability $p(\hat{y}\,|\,h_\mathrm{root})$ of the tree belonging to class $\hat{y}$ via
\begin{equation}
y = \argmax_{\hat{y}} \; p(\hat{y}~|~h_\mathrm{root}) = \argmax_{\hat{y}} \; \mathrm{softmax} \left( W^{(s)} \, h_\mathrm{root} + b^{(s)} \right) ,
\label{equ:softmax}
\end{equation}
with the negative log-likelihood of the true class label $y$ as the cost function \citep{Goodfellow.2017}.

\subsection{Discourse-LSTM}
\label{sec:T_TreeLSTM}

The following section extends the previous Tree-LSTMs through tensor structures. The Discourse-LSTM introduces two modifications that enable us to incorporate (1)~the relation type between two nodes and (2)~the hierarchy type (\ie nucleus or satellite). For this purpose, we replace the usual weight matrices in the tree-structured neural networks with a higher-dimensional representation that allows for additional degrees of freedom with respect to (1) and (2). Thereby, we yield an array of weight matrices, which is formalized and implemented via a tensor.

\begin{figure}
\hspace{0.5cm}
{\includegraphics[height=5cm]{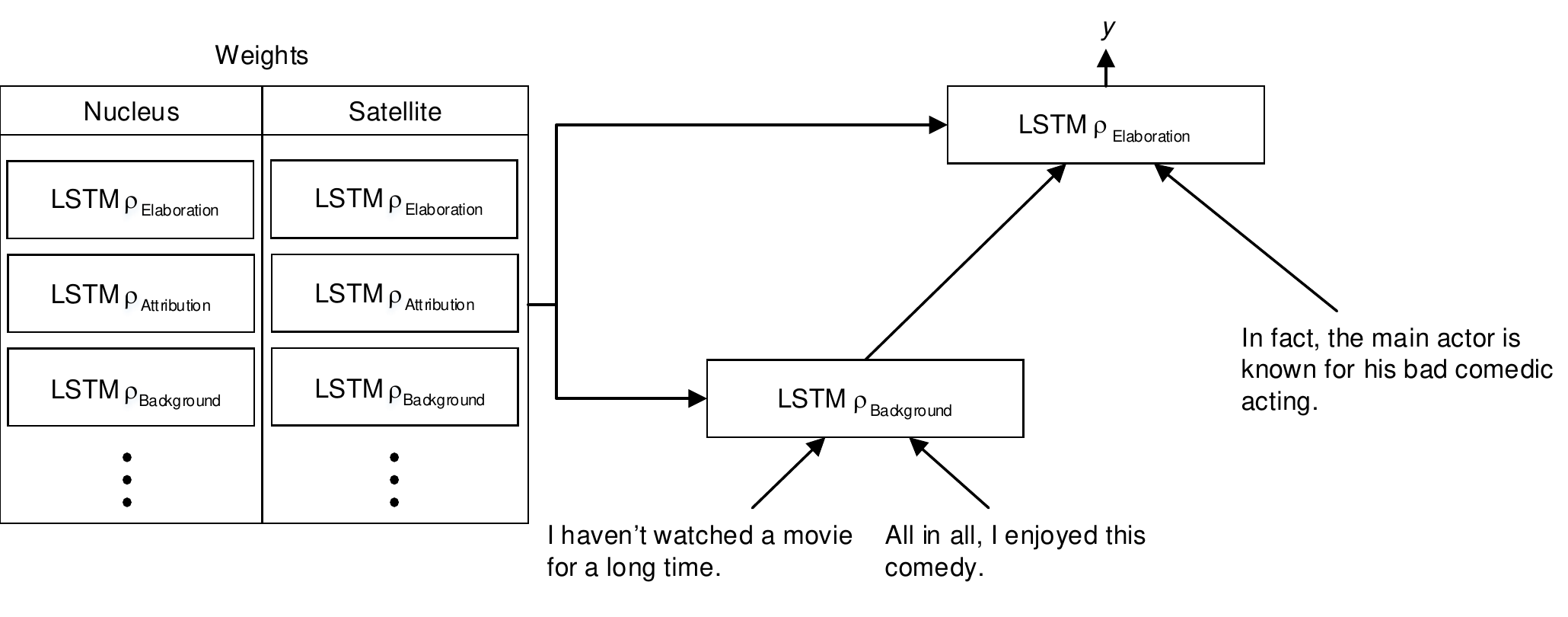}}
\caption{Schematic illustration showing the tensor idea of the Discourse-LSTM. The Discourse-LSTM unit is applied repeatedly at each node in the tree. In contrast to the traditional Tree-LSTM, the Discourse-LSTM stacks multiple LSTM units to account for the relation type between two nodes.}
\label{fig:t-tree-lstm}
\end{figure}

In order to include the relation type, we replace the global LSTM that serves all nodes with one that is dependent on the relation type $r \in \{1,\dots,n\}$. \Cref{fig:t-tree-lstm} visualizes the idea schematically. 
We then select an $\text{LSTM}_{\rho_i}$ for each node depending on its relation type $\rho_i$.

We incorporate the hierarchy type $\tau_i$ (\ie nucleus or satellite) by additionally weighting the cell state $c_j$ and the hidden state $h_j$ before they enter the above tensor-based LSTM. For this purpose, we introduce tensor-based weights 
\begin{align}
\boldsymbol{W}^{(c)} &= \left[W_\mathrm{nucleus}^{(c)};W_\mathrm{satellite}^{(c)}\right],\\
\boldsymbol{W}^{(h)} &= \left[W_\mathrm{nucleus}^{(h)};W_\mathrm{satellite}^{(h)}\right],
\end{align}
where $\boldsymbol{W}^{(c)}$ and $\boldsymbol{W}^{(h)}$ are both of dimensions $2 \times n \times n$, dependent on the input dimension $n$. We then choose the weights according to the hierarchy type $\tau_i$ in the tree. This allows us to additionally discriminate between the influence of nuclei and satellites.

Accordingly, the Discourse-LSTM must simultaneously optimize both the tensor-based LSTM, as well as the hierarchy-related tensors $\boldsymbol{W}^{(c)}$ \emph{and} $\boldsymbol{W}^{(h)}$ based on a combined objective function. We thus rearrange them as \mbox{rank-3} tensors as follows: let $\boldsymbol{W}^{(x)}[r,:]$ indicate the weight tensor for relation type $r$ and $\boldsymbol{W}^{(x)}[l,:]$ denote the weight tensor for a hierarchy type $l \in \left\{ \mathrm{nucleus}, \mathrm{satellite} \right\}$. On this basis, we now specify the new, updated equations for calculating the cell and hidden state. As such, the child-sum Discourse-LSTM computes
\begin{align}
\hat h_k &= \boldsymbol{W}^{(h)}[l,:] \, h_k, \\
\hat c_k &= \boldsymbol{W}^{(c)}[l,:] \, c_k, \\
\tilde h_j &= \sum_{k \in C(j)} \hat{h}_k,\\
i_j &= \mathrm{sigmoid} \left( \boldsymbol{W}^{(i)} x_j + \boldsymbol{U}^{(i)}[r,:] \, \tilde{h}_j + \boldsymbol{b}^{(i)}[r,:] \right),\\
f_{jk} &= \mathrm{sigmoid} \left( \boldsymbol{W}^{(f)} x_j + \boldsymbol{U}^{(f)}[r,:] \, \hat{h}_k + \boldsymbol{b}^{(f)}[r,:] \right), \\
o_j &= \mathrm{sigmoid} \left( \boldsymbol{W}^{(o)} x_j + \boldsymbol{U}^{(o)}[r,:] \, \tilde{h}_j + \boldsymbol{b}^{(o)}[r,:] \right),\\
u_j &= \text{tanh} \left( \boldsymbol{W}^{(u)} x_j + \boldsymbol{U}^{(u)}[r,:] \, \tilde{h}_j +  \boldsymbol{b}^{(u)}[r,:] \right),\\
c_j &= i_j \odot u_j + \sum_{k \in C(j)} f_{jk} \odot \hat c_k,\\
h_j &= o_j \odot \text{tanh}(c_j) .
\end{align}
Similarly, the \mbox{$N$-ary} Discourse-LSTM computes its representations via
\begin{align}
\hat h_{k} &= \boldsymbol{W}^{(h)}[l,:] \, h_{k}, \quad k \in C(j) \\
\hat c_{k} &= \boldsymbol{W}^{(c)}[l,:] \, c_{k}, \quad k \in C(j) \\
i_j &= \mathrm{sigmoid} \left(\boldsymbol{W}^{(i)} x_j + \sum_{m=1}^{N} \boldsymbol{U}_m^{(i)}[r,:] \, \hat{h}_{jm} + \boldsymbol{b}^{(i)}[r,:]\right),\\
f_{jk} &= \mathrm{sigmoid} \left(\boldsymbol{W}^{(f)} x_j + \sum_{m=1}^{N} \boldsymbol{U}_{km}^{(f)}[r,:] \, \hat{h}_{jm} + \boldsymbol{b}^{(f)}[r,:]\right), \\
o_j &= \mathrm{sigmoid} \left(\boldsymbol{W}^{(o)} x_j + \sum_{m=1}^{N} \boldsymbol{U}_m^{(o)}[r,:] \, \hat{h}_{jm} + \boldsymbol{b}^{(o)}[r,:]\right),\\
u_j &= \text{tanh}\left(\boldsymbol{W}^{(u)} x_j + \sum_{m=1}^{N} \boldsymbol{U}_m^{(u)}[r,:] \, \hat{h}_{jm} + \boldsymbol{b}^{(u)}[r,:]\right),\\
c_j &= i_j \odot u_j + \sum_{m=1}^{N} f_{jm} \odot \hat c_{jm},\\
h_j &= o_j \odot \text{tanh}(c_j).
\end{align}
As a result, both the $N$-ary and child-sum Discourse-LSTM integrate the complete discourse tree into the neural network. As opposed to the works in the literature review, this approach allows us to encode both the relation type \emph{and} the hierarchy type.

\subsection{Training data augmentation}

Deep neural networks typically feature a complex structure with thousands of weights that need to be trained, which makes them prone to overfitting. A viable remedy is to artificially increase the number of training samples in order to better tune parameters. Such approaches are common in computer vision, where one extracts different crops from the same image and later considers each as a training instance. We thus propose similar techniques for tree structures  that enlarge our training set. These algorithms take a tree as input and then slightly modify its structure in each epoch of training (one full training cycle on the training set). The first variant, called node reordering, swaps sub-trees, while the second, artificial leaf insertion, randomly exchanges a leaf for a node with two new children. We thereby preserve the tree structure during node reordering, whereas, in artificial leaf insertion, we experiment with how noisy modifications to the tree structure can additionally improve representation learning. 

\subsubsection{Node reordering}
\label{sec:node_reordering}

\begin{figure}
\centering
\makebox[\textwidth]{%
\begin{tabular}{cc}
(a) Tree before node reordering & (b) Tree after node reordering \\
\includegraphics[width=0.5\linewidth]{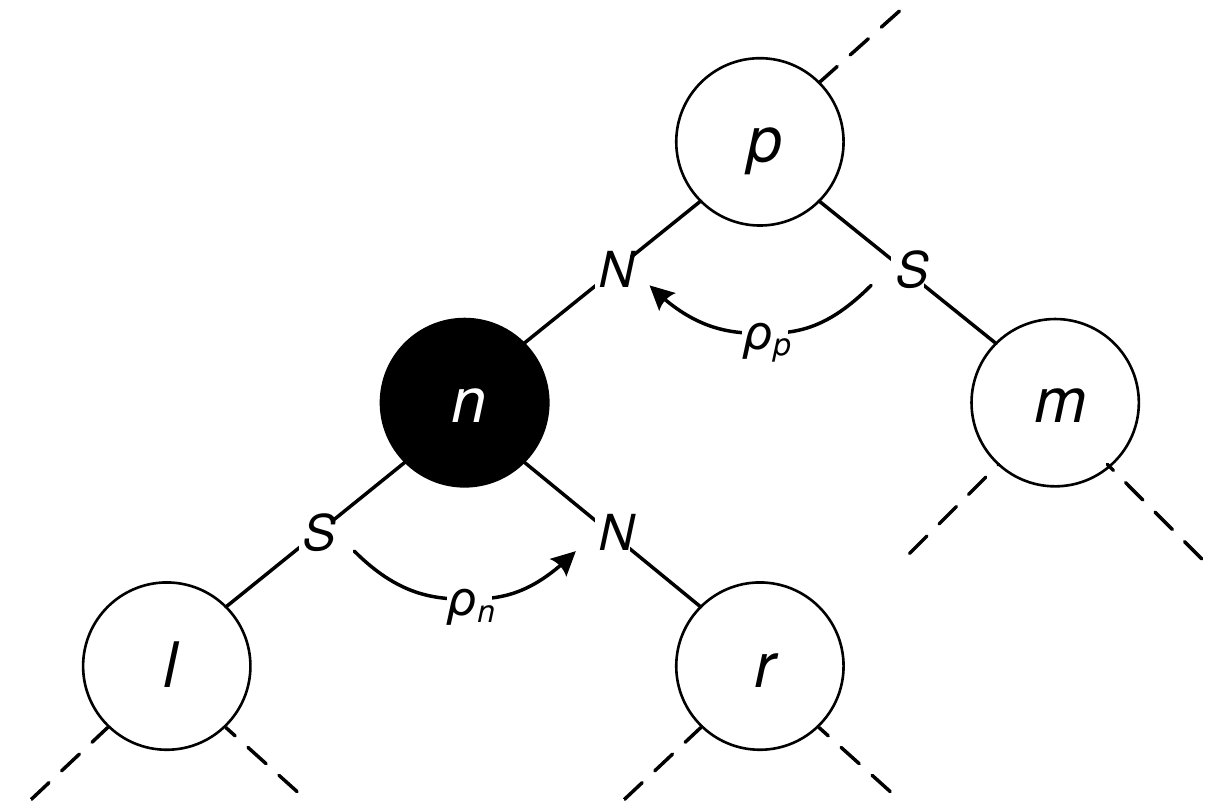} & \includegraphics[width=0.5\linewidth]{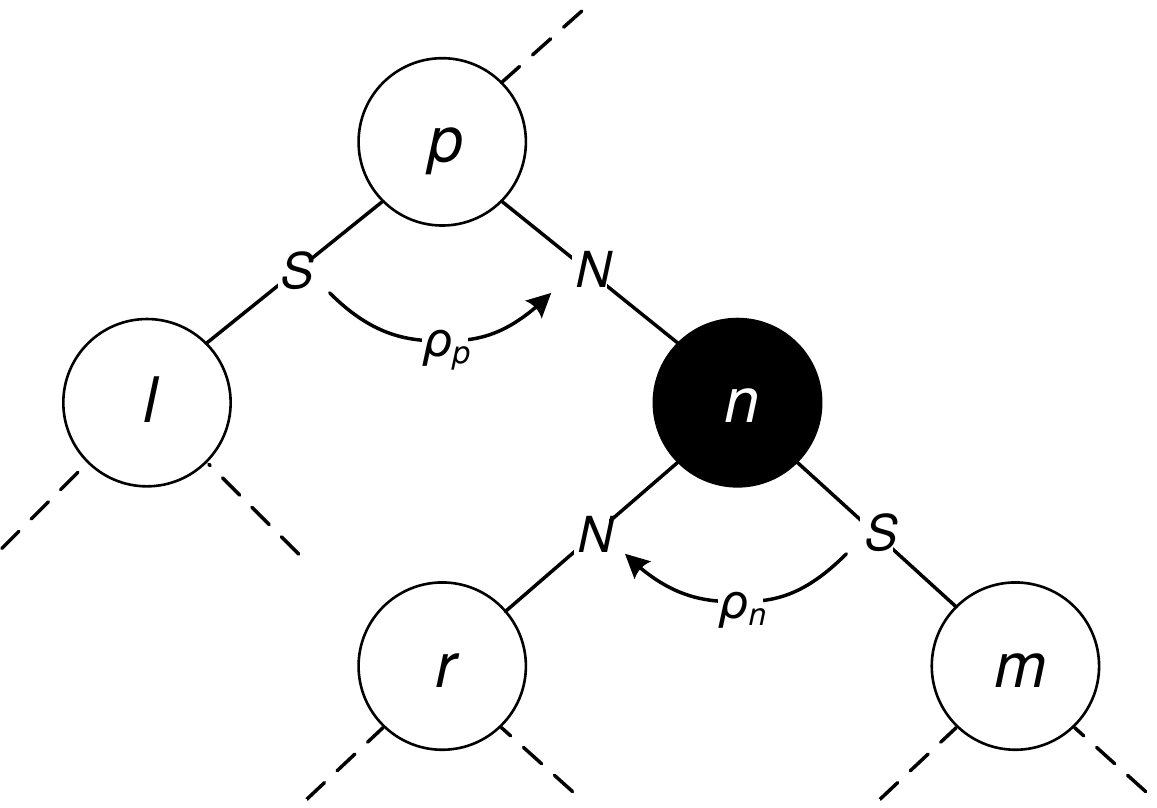} \\
\end{tabular}
}%
\caption{Schematic illustration of node reordering taking place at node $n$. The original tree structure is on the left, while the right shows the tree structure after node reordering.}
\label{fig:node_reordering}
\end{figure}

Node reordering utilizes RST trees and rearranges the positions of inner nodes while trying to preserve the inherent structure. That is, the text passages inside the nodes must maintain their original order since the content might otherwise change its meaning or grammatical structure. Our approach thus randomly chooses an inner node $n$ and relocates it to the position of its sibling $m$ in the tree. Thereby, the position of the sibling is given by the RST structure. The sibling $m$ is then moved down the tree and becomes a child of $n$. Afterwards, the previous position of $n$ is filled by one of its former children. As a result, the order of $l$, $r$ and $m$ from left to right is unchanged. The corresponding algorithm for an inner node $n$ is sketched in \Cref{fig:node_reordering}.

This approach for data augmentation tries to modify the structure slightly, thereby generating potentially different representations of the same tree. The extent of reordering depends on the level of $n$, since a reordering of a node at a higher level usually has a larger effect on the overall tree structure compared to a node at a lower level.

\subsubsection{Artificial leaf insertion}
\label{sec:leaf_insertion}

\begin{figure}\centering
\makebox[\textwidth]{%
\begin{tabular}{cc}
(a) Tree before leaf insertion & (b) Tree after leaf insertion \\
\includegraphics[height=4cm]{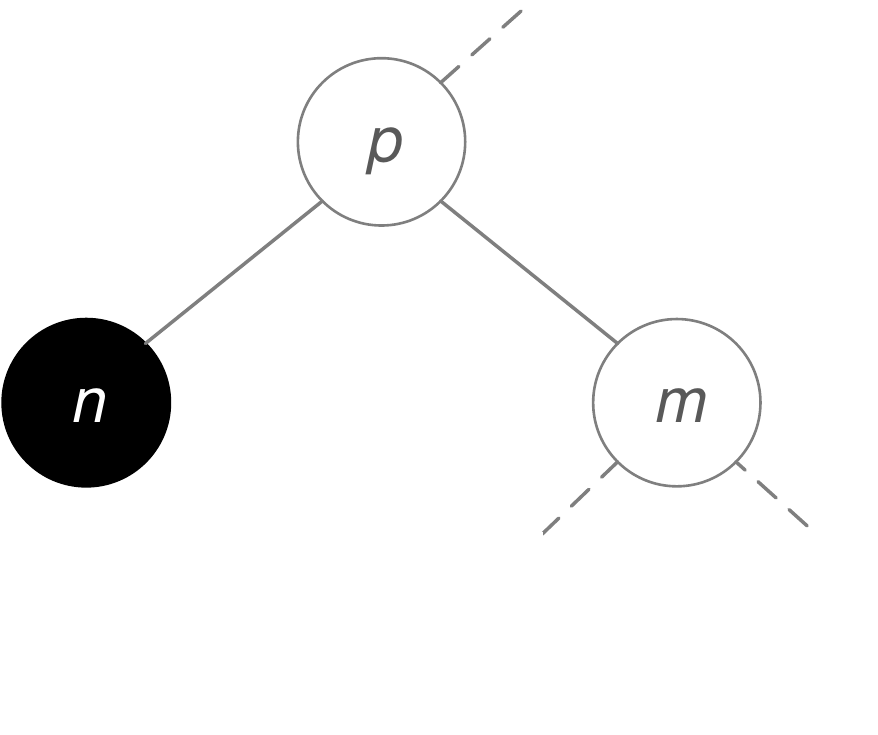} & \includegraphics[height=4cm]{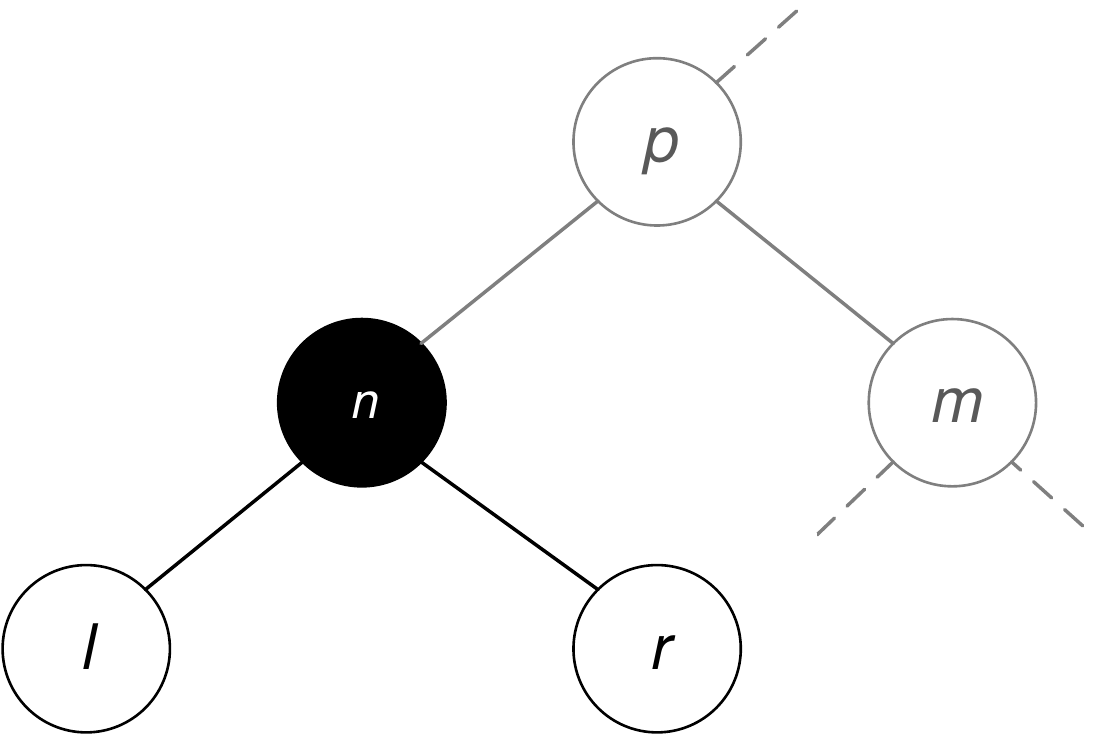} \\
\end{tabular}
}%
\caption{Schematic visualization of artificial leaf insertion at node $n$. The original tree structure is displayed on the left, while the right illustrates the tree structure after artificial leaf insertion.}
\label{fig:Tree_aug_insert}
\end{figure}

Artificial leaf insertion allows us to grow larger trees. Here we make subtle but explicit modifications to the tree structure and hypothesize that, even in presence of the additional noise, this still facilitates representation learning of complex trees. The insertion of leaves into a \mbox{sub-tree} is depicted in \Cref{fig:Tree_aug_insert}. This approach  randomly picks a leaf $n$ from the tree and appends two newly created child nodes $l$ and $r$ which subsequently present the leaves, while $n$ becomes an inner node. We compute $\sigma_l$ and $\sigma_r$ by multiplying $\sigma_n$ with random weights $\omega \in [0,1]$ and $(1-\omega)$, \ie
 \begin{align}
  \sigma_l &= \omega \odot \sigma_n,\\
  \sigma_r &= (1-\omega) \odot \sigma_n.
 \end{align}
These update rules thus try to keep the overall information unchanged, but distribute the values from $n$ into two separate children given a certain ratio $\omega$. We finally choose the relation type $\rho_n$ and the hierarchy type $\tau_n$ randomly. 

\section{Experimental setup}
\label{sec:experiment}

\subsection{Datasets}
\label{sec:dataset}

We build upon earlier work and utilize three common datasets. The first consists of \num{2000} movie reviews from Rotten Tomatoes \citep{Pang.2004}, for which we perform \mbox{\num{10}-fold} \mbox{cross-validation} and then average the predictive performance across splits. The second dataset comprises \num{50000} reviews from the Internet Movie Database~(IMDb), which are split evenly into \num{25000} reviews for training and \num{25000} for testing \citep{Maas.2011}. It includes, at most, \num{30} reviews for any one movie, since reviews for the same movie tend to have correlated ratings. Furthermore, the training and test sets contain a disjoint set of movies to avoid correlation based on movie-specific terms. The third dataset consists of \num{6400} randomly selected food reviews from the Amazon Fine Foods dataset of which \num{3200} are labeled as positive and \num{3200} are labeled as negative \citep{McAuley.2013}. We split the dataset into \num{5120} (\ie \SI{80}{\percent}) reviews for training and \num{1280} (\ie \SI{20}{\percent}) for testing. 

All corpora are preprocessed as follows: we perform tokenization, convert all characters to lowercase, and conduct stemming. The latter maps inflected words onto a base form; \eg \emph{\textquote{enjoyed}} and \emph{\textquote{enjoying}} are both reduced to \emph{\textquote{enjoy}} \citep{Porter.1980}.

\subsection{Descriptive statistics}



The resulting discourse trees exhibit the following characteristics. In the case of reviews from Rotten Tomatoes, they entail \num{51.09} EDUs on average, while this number plummets to \num{19.79} and \num{7.88} EDUs for IMDb reviews and Amazon Food reviews, respectively. The difference stems from the nature of reviews, since Rotten Tomatoes predominantly collects reviews from known critics, while IMDb and Amazon Food reviews are user-generated (and often comprise just a few sentences). The largest discourse tree contains \num{154} levels. \Cref{tbl:rst_relation_desc} reports the relation types and corresponding frequencies in the corpus. The higher number of relations labeled as \emph{elaboration} also has to do with the nature of reviews. Often, the critic presents a thought or argument, which is then followed by further details in support of this claim. When these passages are connected with an additional thought, the EDUs are labeled as a \emph{joint}, thus explaining their overall frequency. The remaining relation types are merely used for specific purposes in the narrative.

\begin{table}
\footnotesize
\centering
\makebox[\textwidth]{%
    \begin{tabular}{l S[table-format=6.0]S S[table-format=7.0]S S[table-format=7.0]S}
		\toprule
		 & 
		\multicolumn{2}{c}{\textbf{\mcellt{Dataset~1:\\ Rotten Tomatoes reviews}}} &
		\multicolumn{2}{c}{\textbf{\mcellt{Dataset~2:\\ IMDb reviews}}} &
		\multicolumn{2}{c}{\textbf{\mcellt{Dataset~3:\\ Food reviews}}} \\
		\cmidrule(l){2-3} \cmidrule(l){4-5} \cmidrule(l){6-7}
		{\textbf{Relation}} & {\textbf{Count}} & {\textbf{Percentage}} & {\textbf{Count}} & {\textbf{Percentage}} & {\textbf{Count}} & {\textbf{Percentage}} \\
		\midrule
		Elaboration & 111330 & \SI{48.24}{\percent} & 677330 & \SI{53.21}{\percent} & 39768 & \SI{48.38}{\percent} \\ 
		Joint  & 58587 & \SI{25.39}{\percent} & 126524 & \SI{9.94}{\percent} & 14883 & \SI{18.11}{\percent} \\
		Attribution  & 13048 & \SI{5.65}{\percent} & 74119 & \SI{5.82}{\percent} & 11204 & \SI{13.63}{\percent} \\
		Textual-organization  & 12788 & \SI{5.54}{\percent} & 235760 & \SI{18.52}{\percent} & 141 & \SI{0.17}{\percent} \\	
		Same-unit  & 12572 & \SI{5.45}{\percent} & 106530 & \SI{8.37}{\percent} & 4461 & \SI{5.43}{\percent} \\	
		Topic-change  & 6950 & \SI{3.01}{\percent} & 3219 & \SI{0.25}{\percent} & 199 & \SI{0.24}{\percent} \\	
		Contrast  & 4404 & \SI{1.91}{\percent} & 7872 & \SI{0.62}{\percent} & 1820 & \SI{2.21}{\percent} \\
		Cause  & 2344 & \SI{1.02}{\percent} & 5467 & \SI{0.43}{\percent} & 1532 & \SI{1.86}{\percent} \\
		Explanation  & 2214 & \SI{0.96}{\percent} & 4070 & \SI{0.32}{\percent} & 602 & \SI{0.73}{\percent} \\
		Condition  & 1951 & \SI{0.85}{\percent} & 19949 & \SI{1.57}{\percent} & 4024 & \SI{4.90}{\percent} \\
		Manner-means  & 1632 & \SI{0.71}{\percent} & 2618 & \SI{0.21}{\percent} & 726 & \SI{0.88}{\percent} \\
		Temporal  & 1382 & \SI{0.60}{\percent} & 5695 & \SI{0.45}{\percent} & 1029 & \SI{1.25}{\percent} \\		
		Comparison  & 854 & \SI{0.37}{\percent} & 3491 & \SI{0.27}{\percent} & 967 & \SI{1.18}{\percent} \\		
		Background  & 346 & \SI{0.15}{\percent} & 5 & \SI{0.00}{\percent} & 111 & \SI{0.14}{\percent} \\
		Topic-comment  & 264 & \SI{0.11}{\percent} & 189 & \SI{0.01}{\percent} & 586 & \SI{0.71}{\percent} \\
		Summary  & 80 & \SI{0.03}{\percent} & 50 & \SI{0.00}{\percent} & 112 & \SI{0.14}{\percent} \\
		Evaluation  & 22 & \SI{0.01}{\percent} & 14 & \SI{0.00}{\percent} & 26 & \SI{0.03}{\percent} \\				
		Enablement  & 0 & \SI{0.00}{\percent} & 3 & \SI{0.00}{\percent} & 2 & \SI{0.00}{\percent} \\	
		\midrule 
		{\textbf{Total}} & 230768 & \SI{100.00}{\percent} & 1272905 & \SI{100.00}{\percent} & 82193 & \SI{100.00}{\percent} \\
    \bottomrule
    \end{tabular}
}
\caption{Descriptive statistics of different relation types in our datasets.}
\label{tbl:rst_relation_desc}
\end{table}

\subsection{Baselines}

We construct na{\"i}ve benchmarks with bag-of-words as follows. We count term frequencies and convert the numerical features into a document-term matrix. As a second baseline, we also scale the term frequencies using the term frequency-inverse document frequency approach~(tf-idf), which puts stronger weights on characteristic terms \citep{Manning.1999}. Both feature spaces are then inserted into a random forest, since this traditional machine learning classifier can detect highly non-linear relationships but still yields satisfactory performance out-of-the-box. These benchmarks allow us to distinguish the sentiment conveyed by words from that conveyed by the discourse structure. 

\subsection{Training process}

We optimize the proposed tree-structured models according to the following process. First, sentiment scores, as well as word embeddings, are fed as leaf node representations into the models. Second, the tree-structured models compute the root node representation which can be utilized for making the prediction through a feedforward layer. Using the prediction along with the label, we then compute the cross-entropy loss made by the model and update the weights with backpropagation. 

\subsection{Model evaluation}


We proceed analogous to \citet{Kraus.2018} in order to tune the model parameters (see appendix). In the case of the random forest baseline, we identify the optimal parameters utilizing a grid search together with \mbox{\num{10}-fold} \mbox{cross-validation} applied to the training set. In contrast, we optimize the deep learning architectures by taking \SI{20}{\percent} of the training data as a validation set. After each epoch, we shuffle the observations and enlarge our training set by constructing additional samples based on our technique for data augmentation. We train our deep learning architecture with early stopping and patience set to ten epochs.


\section{Results}
\label{sec:results}

In this section, we evaluate the performance of our Discourse-LSTM and compare it to the previous baselines. In addition, we perform statistical significance tests on the receiver operating characteristics~(ROC) \citep{DeLong.1988}. The evaluation provides evidence that incorporating semantic structure into the task of sentiment analysis improves the predictive performance. 

\subsection{Dataset 1: movie reviews from Rotten Tomatoes}
\label{sec:rotten_tomatoes_movie_reviews}

\Cref{tbl:RT_Res} details the prediction results for the dataset featuring movie reviews from Rotten Tomatoes. The na{\"i}ve benchmark with tf-idf features yields a balanced accuracy of \num{0.746} and an F1-score of \num{0.763}. The approaches with term frequencies 
 achieve a similar performance. Here we see no clear indication that one of the baselines is consistently superior.

\begin{table}
\scriptsize
\centering
\makebox[\textwidth]{
    \begin{tabular}{ll l SS}
		    \toprule
		    {\textbf{Method}} &
				{\textbf{Variant}} &
		    {\textbf{Data augmentation}} &
		    {\textbf{Balanced accuracy}} &
		    {\textbf{F1-score}} \\
        \midrule
        \multicolumn{5}{l}{\textsc{Benchmark without RST}} \\
        \multicolumn{2}{l}{\quad Sum of all sentiment scores} & -- & 0.609 & 0.640 \\
        \multicolumn{2}{l}{\quad Random forest with term frequency} & -- & 0.762 & 0.752 \\
        \multicolumn{2}{l}{\quad Random forest with tf-idf} & -- & 0.746 & 0.763 \\
        \midrule
        \multicolumn{5}{l}{\textsc{Tree learning with sentiment scores as input}} \\
        \quad Tree-LSTM & Child-sum & -- & 0.780 & 0.772 \\
        \quad Tree-LSTM & Child-sum & Node reordering & 0.785 & 0.778 \\
        \quad Tree-LSTM & Child-sum & Leaf insertion & 0.780 & 0.774 \\
        \quad Tree-LSTM & Child-sum & Node reordering \& leaf insertion & 0.780 & 0.777 \\
        \quad Tree-LSTM & $N$-ary  & -- & 0.770 & 0.768 \\
        \quad Tree-LSTM & $N$-ary & Node reordering & 0.775 & 0.782 \\
        \quad Tree-LSTM & $N$-ary & Leaf insertion & 0.775 & 0.779 \\
        \quad Tree-LSTM & $N$-ary & Node reordering \& leaf insertion & 0.775 & 0.787 \\
        \quad Discourse-LSTM & Child-sum & -- & 0.770 & 0.771 \\
        \quad Discourse-LSTM & Child-sum & Node reordering & \bfseries 0.800 & \bfseries 0.805 \\
        \quad Discourse-LSTM & Child-sum & Leaf insertion & 0.785 & 0.779 \\
        \quad Discourse-LSTM & Child-sum & Node reordering \& leaf insertion & 0.780 & 0.783 \\
        \quad Discourse-LSTM & $N$-ary & -- & 0.770 & 0.772 \\
        \quad Discourse-LSTM & $N$-ary & Node reordering & 0.775 & 0.782 \\
        \quad Discourse-LSTM & $N$-ary & Leaf insertion & 0.775 & 0.787 \\
        \quad Discourse-LSTM & $N$-ary & Node reordering \& leaf insertion & 0.780 & 0.796 \\
        \midrule
        \multicolumn{5}{l}{\textsc{Tree learning with word embeddings as input}} \\
        \quad Tree-LSTM & Child-sum & -- & 0.740 & 0.770 \\
        \quad Tree-LSTM & $N$-ary & -- & 0.695 & 0.738 \\
        \quad Discourse-LSTM & Child-sum & Node reordering \& leaf insertion & 0.550 & 0.521 \\
        \quad Discourse-LSTM & $N$-ary & Node reordering \& leaf insertion & 0.530 & 0.544 \\
        \bottomrule
    \end{tabular}
}
\caption{Predictive performance reported for the test set from the Rotten Tomatoes dataset, including \num{2000} movie reviews.}
\label{tbl:RT_Res}
\end{table}


The simple tree learning based on the Tree-LSTM outperforms all of the previous benchmarks. It achieves a balanced accuracy of up to \num{0.785} and an F1-score of \num{0.787}. 
 Nevertheless, the Tree-LSTM is surpassed by the Discourse-LSTM, which boosts the balanced accuracy to \num{0.800} with an F1-score of \num{0.805}. This amounts to an additional improvement of \num{0.033} (\ie \SI{+4.3}{\percent}) in the F1-score. Altogether, the Discourse-LSTM benefits from the discourse-related information and thus performs best overall. 

Statistical significance tests on the receiver operating characteristics demonstrate that the Discourse-LSTM outperforms the Tree-LSTM to a statistically significant degree at the \SI{1}{\percent} level. Moreover, the child-sum Discourse-LSTM with node reordering improves the predictive performance significantly at the \SI{1}{\percent} level as compared to the child-sum Discourse-LSTM without data augmentation. However, the outcomes are not statistically significant when assessing leaf insertion.

Finally, we additionally note the following patterns: (1)~there is no consistent indication that either the child-sum or $N$-ary variant is consistently superior. (2)~By comparing the underlying algorithms for data augmentation, the results indicate a greater increase in predictive power from node reordering as compared to leaf insertion. This emphasizes that the larger number of training samples outweighs the additional noise from reordering. (3)~The RST-based approaches also outperform models utilizing actual words as features. This suggests that a large portion of sentiment-related information is encoded in the discourse structure. (4)~Utilizing pre-trained word embeddings leads to strong overfitting across all models, thereby lowering the predictive performance. This result stems from the large number of trainable parameters compared to a small number of training samples.


\subsection{Dataset 2: IMDb movie reviews}
\label{sec:imdb_movie_reviews}

\Cref{tbl:IMDB_Res} reports the predictive results for the largest of the three datasets, which is based on \num{50000} IMDb movie reviews. The random forest with tf-idf achieves a performance superior to the previous task, yielding an accuracy of \num{0.825} and an F1-score of \num{0.823}. 

\begin{table}[htbp]
\scriptsize
\centering
\makebox[\textwidth]{
    \begin{tabular}{ll l SS}
		    \toprule
		    {\textbf{Method}} &
				{\textbf{Variant}} &
		    {\textbf{Data augmentation}} & 
		    {\textbf{Balanced accuracy}} &
		    {\textbf{F1-score}} \\
        \midrule
        \multicolumn{5}{l}{\textsc{Benchmarks without RST}} \\
        \multicolumn{2}{l}{\quad Sum of all sentiment scores} & -- & 0.656 & 0.689 \\
        \multicolumn{2}{l}{\quad Random forest with term frequency} & -- & 0.803 & 0.801 \\
        \multicolumn{2}{l}{\quad Random forest with tf-idf} & -- & 0.825 & 0.823 \\
        \midrule
        \multicolumn{5}{l}{\textsc{Tree learning with sentiment scores as input}} \\
        \quad Tree-LSTM & Child-sum & -- & 0.845 & 0.847 \\
        \quad Tree-LSTM & Child-sum & Node reordering & \bfseries 0.850 & 0.848 \\
        \quad Tree-LSTM & Child-sum & Leaf insertion & 0.848 & 0.845 \\
        \quad Tree-LSTM & Child-sum & Node reordering \& leaf insertion & 0.848 & 0.849 \\
        \quad Tree-LSTM & $N$-ary & -- & 0.849 & 0.847 \\
        \quad Tree-LSTM & $N$-ary & Node reordering & \bfseries 0.850 & 0.847 \\
        \quad Tree-LSTM & $N$-ary & Leaf insertion & 0.848 & 0.848 \\
        \quad Tree-LSTM & $N$-ary & Node reordering \& leaf insertion & 0.848 & 0.849 \\
        \quad Discourse-LSTM & Child-sum & -- & \bfseries 0.850 & 0.848 \\
        \quad Discourse-LSTM & Child-sum & Node reordering & \bfseries 0.850 & 0.845 \\
        \quad Discourse-LSTM & Child-sum & Leaf insertion & 0.849 & 0.849 \\
        \quad Discourse-LSTM & Child-sum & Node reordering \& leaf insertion & \bfseries 0.850 & 0.847 \\
        \quad Discourse-LSTM & $N$-ary & -- & 0.849 & 0.847 \\
        \quad Discourse-LSTM & $N$-ary & Node reordering & \bfseries 0.850 & 0.849 \\
        \quad Discourse-LSTM & $N$-ary & Leaf insertion & \bfseries 0.850 & 0.846 \\
        \quad Discourse-LSTM & $N$-ary & Node reordering \& leaf insertion & 0.848 & 0.848 \\
        \midrule
        \multicolumn{5}{l}{\textsc{Tree learning with word embeddings as input}} \\
        \quad Tree-LSTM & Child-sum & -- & 0.849 & 0.848 \\
        \quad Tree-LSTM & $N$-ary & -- & 0.847 & 0.849 \\
        \quad Discourse-LSTM & Child-sum & Node reordering \& leaf insertion & 0.671 & 0.562 \\
        \quad Discourse-LSTM & $N$-ary & Node reordering \& leaf insertion & 0.847 & \bfseries 0.852 \\
        \bottomrule
    \end{tabular}
}
\caption{Predictive performance reported for the test set from the IMDb dataset.}
\label{tbl:IMDB_Res}
\end{table}


Tree-structured LSTMs outperform our baseline models. For instance, the $N$-ary Tree-LSTM raises the balanced accuracy and the F1-score of the na{\"i}ve baselines by \num{0.025} and \num{0.026}, respectively. Our Discourse-LSTMs achieve a similar balanced accuracy of \num{0.850} compared to simple Tree-LSTMs; however, results of the Discourse-LSTMs are more consistent. It achieves an accuracy of \num{0.850} and an F1-score of \num{0.849} by utilizing data augmentation. Pre-trained word embeddings push the F1-score of the $N$-ary Discourse-LSTM with data augmentation to \num{0.852}. Again, we find no general pattern indicating that one technique for enlarging the training set scores better than the other. 

Statistical tests show that the $N$-ary Discourse-LSTM with node reordering performs significantly better than the Tree-LSTM at the \SI{10}{\percent} level. Also, the $N$-ary Discourse-LSTM with node reordering performs significantly better at the \SI{10}{\percent} level as compared to the $N$-ary Discourse-LSTM without data augmentation.

\subsection{Dataset 3: Amazon Fine Food reviews}

\Cref{tbl:Food_Res} lists the prediction results for the dataset featuring food reviews left by Amazon users. Regarding traditional machine learning, the random forest with tf-idf features achieves a balanced accuracy of \num{0.742} and an F1-score of \num{0.770}. 

\begin{table}[htbp]
\scriptsize
\centering
\makebox[\textwidth]{
    \begin{tabular}{ll l SS}
		    \toprule
		    {\textbf{Method}} &
				{\textbf{Variant}} &
		    {\textbf{Data augmentation}} & 
		    {\textbf{Balanced accuracy}} &
		    {\textbf{F1-score}} \\
        \midrule
        \multicolumn{5}{l}{\textsc{Benchmarks without RST}} \\
        \multicolumn{2}{l}{\quad Sum of all sentiment scores} & -- & 0.662 & 0.725 \\
        \multicolumn{2}{l}{\quad Random forest with term frequency} & -- & 0.713 & 0.750 \\
        \multicolumn{2}{l}{\quad Random forest with tf-idf} & -- & 0.742 & 0.770 \\
        \midrule
        \multicolumn{5}{l}{\textsc{Tree learning with sentiment scores as input}} \\
        \quad Tree-LSTM & Child-sum & -- & 0.801 & 0.791 \\
        \quad Tree-LSTM & Child-sum & Node reordering & 0.805 & 0.793 \\
        \quad Tree-LSTM & Child-sum & Leaf insertion & 0.805 & 0.791 \\
        \quad Tree-LSTM & Child-sum & Node reordering \& leaf insertion & 0.806 & 0.793 \\
        \quad Tree-LSTM & $N$-ary & -- & 0.805 & 0.787 \\
        \quad Tree-LSTM & $N$-ary & Node reordering & 0.807 & 0.789 \\
        \quad Tree-LSTM & $N$-ary & Leaf insertion & 0.808 & 0.791 \\
        \quad Tree-LSTM & $N$-ary & Node reordering \& leaf insertion & 0.807 & 0.789 \\
        \quad Discourse-LSTM & Child-sum & -- & 0.806 & 0.789 \\
        \quad Discourse-LSTM & Child-sum & Node reordering & 0.808 & 0.793 \\
        \quad Discourse-LSTM & Child-sum & Leaf insertion & 0.808 & 0.791 \\
        \quad Discourse-LSTM & Child-sum & Node reordering \& leaf insertion & 0.807 & 0.789 \\
        \quad Discourse-LSTM & $N$-ary & -- & 0.811 & 0.802 \\
        \quad Discourse-LSTM & $N$-ary & Node reordering & 0.812 & \bfseries 0.803 \\
        \quad Discourse-LSTM & $N$-ary & Leaf insertion & 0.804 & 0.790 \\
        \quad Discourse-LSTM & $N$-ary & Node reordering \& leaf insertion & \bfseries 0.813 & 0.801 \\
        \midrule
        \multicolumn{5}{l}{\textsc{Tree learning with word embeddings as input}} \\
        \quad Tree-LSTM & Child-sum & -- & 0.760 & 0.747 \\
        \quad Tree-LSTM & $N$-ary & -- & 0.771 & 0.762 \\
        \quad Discourse-LSTM & Child-sum & Node reordering \& leaf insertion & 0.737 & 0.739 \\
        \quad Discourse-LSTM & $N$-ary & Node reordering \& leaf insertion & 0.525 & 0.448 \\
        \bottomrule
    \end{tabular}
}
\caption{Predictive performance reported for the test set from the Amazon Fine Food dataset, with \num{6400} randomly picked reviews.}
\label{tbl:Food_Res}
\end{table}

Tree-LSTMs outperform all baselines with term frequency features.
 For instance, the $N$-ary Tree-LSTM leads to a balanced accuracy of \num{0.805} and an F1-score of \num{0.787}. When exploiting all information from the RST tree, the balanced accuracy increases further to \num{0.813}, along with an F1-score of \num{0.801}. Therefore, data augmentation leveraged the balanced accuracy by \num{0.002} but decreased the F1-score by \num{0.001}. Tree-structured models utilizing pre-trained word embeddings outperform the random forest with both tf and tf-idf features, showing a balanced accuracy of \num{0.771} and an F1-score of \num{0.762}. However, word embeddings yield inferior performance as compared to the Tree-LSTM and Discourse-LSTM with sentiment scores.

Statistical tests on the ROC curves show that the performance of the $N$-ary Discourse-LSTM is significantly better compared to both Tree-LSTMs at the \SI{1}{\percent} level. Although the $N$-ary Discourse LSTM benefits from node reordering, showing a higher balanced accuracy and F1-score, the improvement is not significant.

\subsection{Comparison}

In the following, we compare our Discourse-LSTM to the relation-specific approach in \citet{Ji.2017}. In contrast to ours, it sums the representations in each recursive cell and thus cannot distinguish between nucleus and satellite. In addition, their approach utilizes a recursive neural network, which is known to suffer from vanishing or exploding gradients \citep{Bengio.1994}. In response to such shortcomings, we decided to utilize a long short-term memory. 

We proceed as follows in order to specifically compare their approach to ours. We leave all other parameters unchanged (\ie identical to the previous experiments). We thus feed the networks with EDU-level features from the previous dictionary-based sentiment scores. The performance measurements indicate that the resulting predictive accuracy is inferior to the Discourse-LSTM. For the dataset from Rotten Tomatoes, their approach achieves a balanced accuracy of \num{0.775} and thus represents a decline of \num{0.025} (\ie \SI{-3.2}{\percent}) compared to the best-performing child-sum Discourse-LSTM. In the case of the IMDb reviews and Amazon Fine Food reviews, their approach yields a balanced accuracy of \num{0.831} and \num{0.803}, while the Discourse-LSTM achieves \num{0.850} and \num{0.813}, respectively. Hence, this work results in an improvement of \num{0.019} (\ie \SI{+2.3}{\percent}) and \num{0.010} (\ie \SI{+1.2}{\percent}).


We additionally compare our proposed methodology for diminishing the effect of overfitting against the widely-utilized dropout technique. Dropout, in contrast to our approach of data augmentation, reduces overfitting by randomly dropping out a certain share of neurons in order to improve generalizability of the network. This prevents the neurons from co-adapting too much during training \citep{Srivastava.2014}.

In order to compare dropout to node reordering and leaf insertion, we perform the following experiment utilizing the Rotten Tomatoes dataset with the $N$-ary Discourse-LSTM and the Child-sum Discourse-LSTM. While training the models, we randomly choose a certain share of weights that is set to zero. Thereby, the set of dropped-out neurons changes in each iteration of training and is defined by a dropout probability $p$, \ie the probability of a random weight being set to zero. In this comparison, we experiment with $p$ set to \num{0.1}, \num{0.2}, \num{0.5} and \num{0.7}. For the $N$-ary Discourse-LSTM, dropout increases the balanced accuracy by \num{0.015} (\ie \SI{+1.9}{\percent}), whereas data augmentation increases the balanced accuracy by \num{0.01} (\ie \SI{+1.3}{\percent}). However, for the Child-sum Discourse-LSTM, data augmentation leads to a greater improvement of \num{0.03} (\ie \SI{+3.9}{\percent}) compared to the improvement of \num{0.02} (\ie \SI{+2.6}{\percent}) when utilizing dropout.

In a further experiment, we combine dropout, node reordering and leaf insertion in order to examine the universal applicability of our approach. In this experiment, we see an improvement of \num{0.03} (\ie \SI{+3.9}{\percent}) for the Child-sum Discourse-LSTM, which is on par with the results we obtained when utilizing data augmentation alone. Yet, for the $N$-ary Discourse-LSTM, we see a performance increase of \num{0.02} (\ie \SI{+2.6}{\percent}) when utilizing dropout, node reordering and leaf insertion together. Thus, this combination outperforms models that only use a single method to avoid overfitting.

\subsection{Sensitivity analysis}

We now investigate the sensitivity of our models to the quality of the RST parser. Therefore, we replace a varying percentage of relation types with random noise. Finally, we evaluate the performance of a $N$-ary Discourse-LSTM with noisy data and compare it to the performance on the original trees. For this analysis, we utilize the Amazon Fine Food dataset.

\Cref{tbl:Sens_hier_types} shows the results. When modifying \SI{2}{\percent} of the relation types, we see no difference in terms of balanced accuracy, but a decrease of \num{0.007} points in the F1-score. By altering \SI{10}{\percent} of the relation types, the balanced accuracy decreases to \num{0.803} with an F1-score of \num{0.794}. This reduction is statistically significant at the \SI{1}{\percent} level. When modifying \SI{20}{\percent} of the relation types, the performance decreases further to a balanced accuracy of \num{0.794} and an F1-score of \num{0.788}. 

\begin{table}
\scriptsize
\centering
\makebox[\textwidth]{
    \begin{tabular}{r SS}
		    \toprule
		    {\textbf{Percentage of noisy}} &
		    {\textbf{Balanced accuracy}} &
		    {\textbf{F1-score}} \\
		    {\textbf{hierarchy types}} \\
        \midrule
        None & 0.811 & 0.802 \\
        \midrule
        1\,\% & 0.809 & 0.797 \\
        2\,\% & 0.811 & 0.795 \\
        5\,\% & 0.809 & 0.800 \\
        10\,\% & 0.803 & 0.794 \\
        20\,\% & 0.794 & 0.788 \\
        \bottomrule
    \end{tabular}
}
\caption{Sensitivity analysis of how the discourse parser affects the predictive performance of our Discourse-LSTM. Here a varying proportion of relation types are replaced by random noise. The results originate from the Amazon Fine Food dataset and an $N$-ary Discourse-LSTM.}
\label{tbl:Sens_hier_types}
\end{table}

\subsection{Discussion}
\label{sec:discussion}

\begin{figure}
\hspace{-0.2cm}
\includegraphics[width=1.0\textwidth]{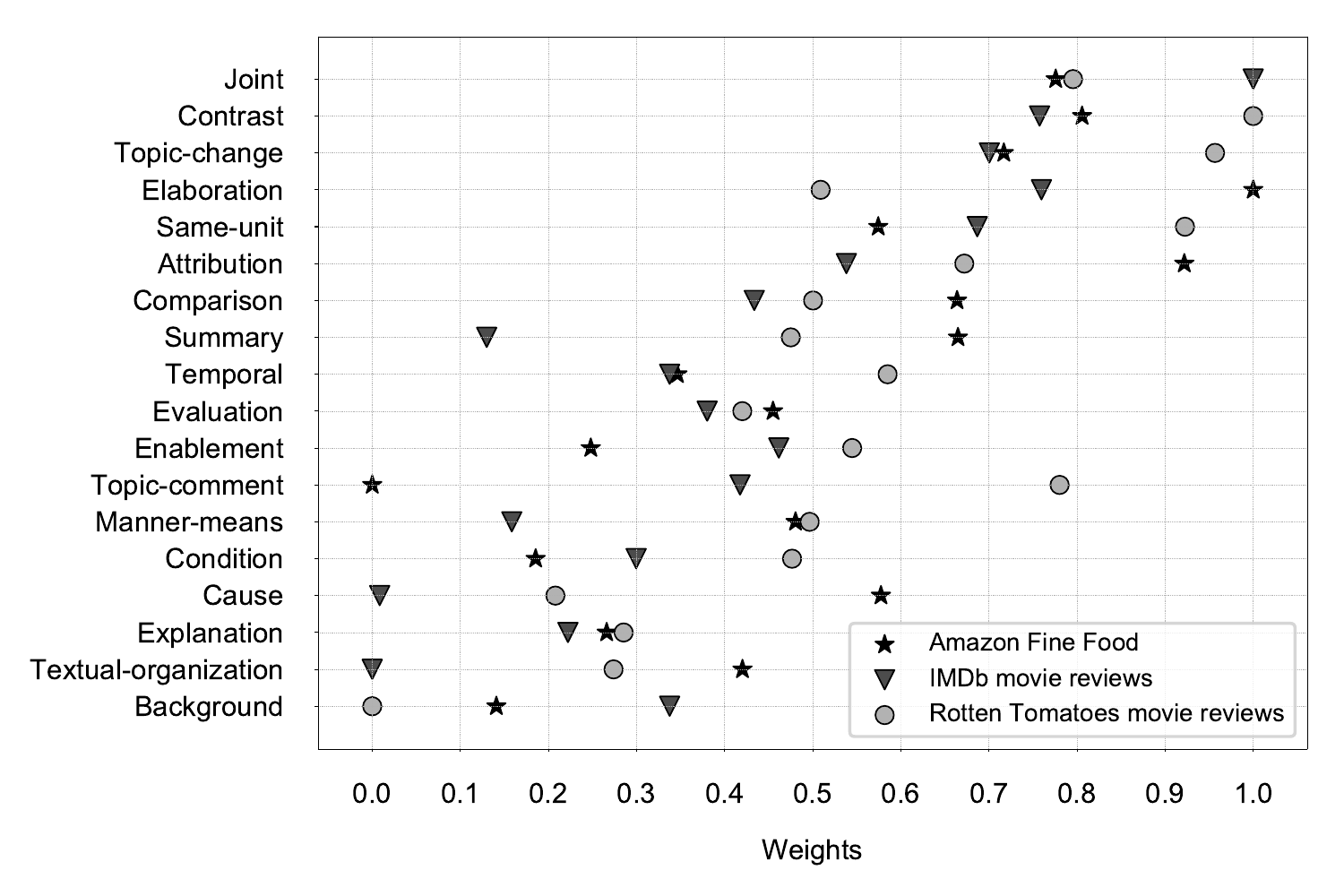}
\caption{The weights from the tensors reveal the relative importance of the discourse. More precisely, the plot shows the normalized weights of the tensor $\boldsymbol{U}^{(u)}_m$ across different relation types $m$. The results stem from the child-sum Discourse-LSTM without data augmentation.}
\label{fig:tensorweights}
\end{figure}

We now investigate the trained weights of our tensor-based mechanism inside the Discourse-LSTM. This facilitates insights into how the neural network processes the discourse and infers the sentiment from the semantic structure of textual materials. \Cref{fig:tensorweights} compares the normalized weights of the tensors $\boldsymbol{U}^{(u)}_m$ across different relation types $m$. The values result from using a child-sum Discourse-LSTM without data augmentation. Overall, the tensor weights between both datasets are highly correlated. For instance, the correlation coefficient between IMDb and Rotten Tomatoes stands at \num{0.640}, statistically significant at the \SI{1}{\percent} level. However, we observe large differences in the relative importance across the relation types. For instance, relation types such as \emph{background} and \emph{textual-organization} entail only marginal importance, consistent with initial expectations. In contrast, the \emph{joint} relation yields among the highest weights across both datasets. 

With regard to the hierarchy-related tensors, we find a greater importance (\ie higher weights) for nuclei as compared to satellites. For instance, the IMDb movie reviews lead to a nucleus weight of \num{0.738}, whereas the weight of satellites totals a mere \num{0.588}. This is in line with our intuition and the idea of RST: nuclei are supposed to be more essential to the writer's purpose than satellites.

\Cref{fig:sample_text} shows the obtained results with an illustrative example. Here we color the text according to the tensor values inside the child-sum Discourse-LSTM without data augmentation. A red text color refers to more essential pieces of information as compared to blue. In example~(a), the Discourse-LSTM assigns the highest relevance to the passage \emph{\textquote{All in all, I enjoyed this comedy}}, whereas it gives the least emphasis to \emph{\textquote{I haven't watched a movie for a long time}}. In example~(b) from Rotten Tomatoes, the Discourse-LSTM gives highest weight to the passage \emph{\textquote{Kolya is one of the richest films i've seen in some time}}. It assigns the lowest relevance to the second to fourth passage, which describe the plot of the movie. In example~(c) the Discourse-LSTM gives highest weight to the passages \emph{\textquote{it will be a stinker, and to everybody's surprise ( perhaps even the studio ) the film becomes a critical darling.}} and \emph{\textquote{The plot is deceptively simple}}. 

\begin{figure}
\begin{subfigure}{1.0\textwidth}
\subcaption{Short example.}
\fbox{
\begin{minipage}[t]{1.0\textwidth}
\textcolor{red!0!blue}{I haven't watched a movie for a long time.} $\big \vert$ 
\textcolor{red!40!red}{All in all, I enjoyed this comedy.} $\big \vert$ 
\textcolor{red!40!blue}{In fact, the main actor is known for is bad comedic acting.}
\end{minipage}
}
\end{subfigure}


\begin{subfigure}{1.0\textwidth}
\subcaption{Medium-length review.}
\fbox{
\begin{minipage}{1.0\textwidth}
\textcolor{red!40!red}{Kolya is one of the richest films i've seen in some time.} $\big \vert$ 
\textcolor{red!0!blue}{Zdenek Sverak plays a confirmed old bachelor,} $\big \vert$ 
\textcolor{red!0!blue}{who finds his life as a czech cellist increasingly impacted by the five-year old boy} $\big \vert$ 
\textcolor{red!0!blue}{that he's taking care of.} $\big \vert$ 
\textcolor{red!40!blue}{Though it ends rather abruptly} $\big \vert$ 
\textcolor{red!40!blue}{-- and i'm whining,} $\big \vert$ 
\textcolor{red!40!blue}{'cause i wanted to spend more time with these characters --} $\big \vert$ 
\textcolor{red!40!blue}{the acting, writing, and production values are as high as,} $\big \vert$  
\textcolor{red!40!blue}{if not higher than,} $\big \vert$ 
\textcolor{red!40!blue}{comparable american dramas.} $\big \vert$ 
\end{minipage}
}
\end{subfigure}


\begin{subfigure}{1.0\textwidth}
\subcaption{Long review}
\fbox{
\begin{minipage}{1.0\textwidth}
\textcolor{red!40!blue}{Every now and then a movie comes along from a suspect studio, with every indication that} $\big \vert$ 
\textcolor{red!40!red}{it will be a stinker,} $\big \vert$ 
\textcolor{red!40!red}{and to everybody's surprise ( perhaps even the studio ) the film becomes a critical darling.} $\big \vert$ 
\textcolor{red!0!blue}{MTV films' election , a high school comedy} $\big \vert$ 
\textcolor{red!40!blue}{starring Matthew Broderick and Reese Witherspoon, is a current example.} $\big \vert$
\textcolor{red!0!blue}{Did anybody know this film existed a week before it opened?} $\big \vert$
\textcolor{red!40!red}{The plot is deceptively simple.} $\big \vert$
\textcolor{red!0!blue}{George Washington carver high school is having student elections.} $\big \vert$ 
\textcolor{red!0!blue}{Tracy flick ( Reese Witherspoon ) is an over-achiever with her hand} $\big \vert$ 
\textcolor{red!0!blue}{raised at nearly every question , way , way , high.} $\big \vert$
\textcolor{red!0!blue}{Mr.} $\big \vert$ 
\textcolor{red!0!blue}{" m " ( Matthew Broderick ), sick of the megalomaniac student, encourages paul,} $\big \vert$
\textcolor{red!0!blue}{a popular-but-slow jock to run.} $\big \vert$
\textcolor{red!40!blue}{And Paul's nihilistic sister} $\big \vert$ 
\textcolor{red!40!blue}{jumps in the race as well, for personal reasons \dots} 
\end{minipage}
}
\end{subfigure}
\caption{Three illustrative examples after pre-processing. Individual EDUs are separated by vertical bars. A red text color highlights more relevant passages as measured by a higher weight of the tensor $\boldsymbol{U}^{(u)}_m$ inside the Discourse-LSTM. Purple and blue text highlights little relevant and irrelevant passages.}
\label{fig:sample_text}
\end{figure}


The above discussion confirms that the tensors build a mechanism that learns to weight the importance of sentences based on their position and relations in the discourse tree. As a result, the Discourse-LSTM can localize the relevant parts of the document and ascertain the relative importance of sentiment scores

\section{Conclusion}
\label{sec:conclusion}

Deep learning for natural language predominantly builds upon sequential models such as LSTMs. 
 While these models usually achieve a high predictive power when applied to short texts, the complexity of linguistic discourse hampers performance for longer documents. As a remedy, our paper proposes an innovative, discourse-aware approach: we first parse the semantic structure based on rhetorical structure theory, thereby mapping the document onto a discourse tree that encodes its storyline. We then apply tailored tree-structured deep neural networks with an additional tensor structure that enables us to directly learn the complete discourse tree. Each of the architectures entails more than \num{10000} parameters, empowering the models to learn highly non-linear relationships.

Our findings reveal that our Discourse-LSTM substantially outperforms the baselines. For instance, the best-performing Discourse-LSTMs achieve improvements of \SI{4.27}{\percent} (Rotten Tomatoes), \SI{0.60}{\percent} (IMDb) and \SI{1.52}{\percent} (Amazon Fine Food reviews) in the F1-score as compared to using simple Tree-LSTMs. These gains are partially a result of our techniques for data augmentation, which slightly alter existing trees in order to enlarge the size of the training set. Evidently, data augmentation presents a viable option to reduce the risk of overfitting. Furthermore, the underlying tensor structure learns the relative importance of passages based on their position in the discourse tree. This facilitates insights into which discourse units convey essential pieces of information. 


\appendix
\section{Tuning ranges}

\begin{table}[H]
\footnotesize
\begin{tabular}{lll}
\toprule
\textbf{Predictive model} & \textbf{Parameter} & \textbf{Tuning range} \\ 
\midrule
Random forest & Number of randomly-sampled variables & 1, 2, 3, 5, 7 \\ 
& Number of trees & 100, 200, 500, 1000 \\ 
& Maximum depth of trees & 1, 5, 10, 20, 50, 100, Unconstrained \\[0.5em]
Tree-LSTM & Regularization strength & 0.001, 0.01, 0.05 \\
& Input dimension for sentiment scores & 1 \\
& Input dimension for word embeddings & 50 \\
& Memory size & 10, 20, 50 \\
& Learning rate & 0.00001, 0.0001, 0.001 \\[0.5em]
Discourse-LSTM & Input dimension for sentiment scores & 1 \\
& Input dimension for word embeddings & 50 \\
& Memory size & 10, 20, 50 \\
& Regularization strength & 0.001, 0.01, 0.05 \\
& Learning rate & 0.00001, 0.0001, 0.001 \\
\bottomrule
\end{tabular}
\caption{Overview of model parameters and their tuning ranges.}
\label{tbl:tuning}
\end{table} 

\section*{Acknowledgment}

{\footnotesize 
The valuable contribution of Ryan Grabowski is greatly appreciated.
}



\bibliographystyle{model5-names}
\bibliography{literature}







\end{document}


\pagenumbering{gobble}%
\appendix

\section{Preprocessing}

Before performing the actual sentiment analysis, there are several preprocessing steps as follows:
\begin{enumerate}
\item \textbf{Filtering.} We apply the following filter rules: first, each announcement must have at least 50~words. Second, we focus only on ad~hoc press releases from German companies which are written in the English language. Our final corpus consists of 14,427 ad~hoc announcements. To study stock market reaction, we use the daily stock market returns of the corresponding company, originating from Thomson Reuters Datastream. We only include business days and take the first message of each day; yielding a total of 1,892~observations. In addition, we adjust the publication days of ad~hoc announcements according to the opening times of the stock exchange. This is achieved by assigning all disclosures filed after 8~p.\,m. to the next day. 
\item \textbf{Tokenization.} Corpus entries are split into single words named \emph{tokens}~\cite{Grefenstette.1994}. 
\item \textbf{Negations.} Negations invert the meaning of words and sentences~\cite{Dadvar.2011,Prollochs.2015,Negations.DSS}. When encountering the word \emph{no}, each of the subsequent three words (\ie the object) are counted as words from the opposite dictionary. When other negating terms are encountered (\emph{rather}, \emph{hardly}, \emph{couldn't}, \emph{wasn't}, \emph{didn't}, \emph{wouldn't}, \emph{shouldn't}, \emph{weren't}, \emph{don't}, \emph{doesn't}, \emph{haven't}, \emph{hasn't}, \emph{won't}, \emph{hadn't}, \emph{never}), the meaning of all succeeding words is inverted.
\item \textbf{Stop word removal.} Words without a deeper meaning, such as \emph{the}, \emph{is}, \emph{of}, etc. are named \emph{stop words}~\cite{Manning.1999} and can be removed. We use a list of 571~stop words proposed in~\cite{Lewis.2004}.
\item \textbf{Synonym merging.} Synonyms, though spelled differently, convey the same meaning. In order to group synonyms by their meaning, we follow a method that is referred to as pseudoword generation~\cite{Manning.1999}. \Cref{tbl:pseudowords} provides list with approximately 175~frequent synonyms or phrases from the finance domain, which we utilize to aggregate synomyms according to their meanings. In the case of \eg \emph{at least}, we use pseudoword generation to map a common phrase with more than one word onto a single token.
\item \textbf{Stemming.} Stemming refers to the process of reducing inflected words to their stem~\cite{Manning.1999}. Here, we use the so-called Porter stemming algorithm~\cite{Porter.1980}. 
\end{enumerate}

\section*{References}



\bibliographystyle{model1-num-names}
\bibliography{literature}





